\pdfoutput=1

\documentclass[11pt]{article}

\usepackage[final]{acl}

\usepackage{times}
\usepackage{latexsym}
\usepackage{afterpage}
\usepackage{booktabs}
\usepackage{graphicx}
\usepackage{threeparttable}
\usepackage{microtype}

\newsavebox{\myentiretablebox}

\usepackage[T1]{fontenc}
\usepackage[utf8]{inputenc}
\usepackage{microtype}
\usepackage{inconsolata}
\usepackage{graphicx}
\usepackage{float}
\usepackage{subcaption}
\usepackage{adjustbox}
\usepackage{multirow}
\usepackage{wrapfig}
\usepackage{booktabs}
\usepackage{xurl} 
\usepackage{hyperref}   
\hypersetup{breaklinks=true} 
\hypersetup{colorlinks,linkcolor={red}} \usepackage{url}
\usepackage{pifont}
\usepackage{enumitem}
\usepackage{caption}
\usepackage{amsmath}
\usepackage{tabularray}

\title{Task-Aware Resolution Optimization for Visual Large Language Models}

\author{
  \textbf{Weiqing Luo}\textsuperscript{\ding{171},\ding{169}} \thanks{The work is done as an intern at UNC Chapel Hill.}\quad 
  \textbf{Zhen Tan}\textsuperscript{\ding{169}} \quad
  \textbf{Yifan Li} \textsuperscript{\ding{168}} \quad
  \textbf{Xinyu Zhao}\textsuperscript{\ding{171}} \quad \\
  \textbf{Kwonjoon Lee}\textsuperscript{\ding{170}} \quad
  \textbf{Behzad Dariush} \textsuperscript{\ding{170}} \quad
  \textbf{Tianlong Chen}\textsuperscript{\ding{171}} \\ 
  \textsuperscript{\ding{171}}University of North Carolina at Chapel Hill \quad
  \textsuperscript{\ding{169}}Arizona State University \\
  \textsuperscript{\ding{168}}Michigan State University \quad
  \textsuperscript{\ding{170}}Honda Research Institute USA\\
  {\tt \{weiqing2,ztan36\}@asu.edu} \quad
  {\tt liyifa11@msu.edu}\\
  {\tt \{kwonjoon\_lee,bdariush\}@honda-ri.com} \quad
  {\tt \{xinyu,tianlong\}@cs.unc.edu}\\
}

\definecolor{mylightgray}{gray}{0.9}
\definecolor{mediumgray}{gray}{0.5}
\newcommand{\graytext}[1]{\textcolor{mediumgray}{#1}}
\definecolor{lightgraytext}{gray}{0.55} 

\newcommand*{\SuperScriptSameStyle}[1]{%
  \ensuremath{%
    \mathchoice
      {{}^{\displaystyle #1}}%
      {{}^{\textstyle #1}}%
      {{}^{\scriptstyle #1}}%
      {{}^{\scriptscriptstyle #1}}%
  }%
}
\newcommand*{\oneS}{\SuperScriptSameStyle{*}}

\begin{document}

\setlength{\abovedisplayskip}{1pt}
\setlength{\belowdisplayskip}{1pt}

\maketitle
\footnotetext{Accepted as a main conference paper at the 2025 Conference on Empirical Methods in Natural Language Processing (EMNLP 2025).}
\begin{abstract}
Real-world vision-language applications demand varying levels of perceptual granularity. However, most existing visual large language models (VLLMs), such as LLaVA, pre-assume a fixed resolution for downstream tasks, which leads to subpar performance. To address this problem, we \underline{\textit{first}} conduct a comprehensive and pioneering investigation into the resolution preferences of different vision-language tasks, revealing a correlation between resolution preferences with \ding{182}~image complexity, and \ding{183}~uncertainty variance of the VLLM at different image input resolutions. Building on this insight, we propose an empirical formula to determine the optimal resolution for a given vision-language task, combining these two factors. \underline{\textit{Second}}, based on rigorous experiments, we propose a novel parameter-efficient fine-tuning technique to extend the visual input resolution of pre-trained VLLMs to the identified optimal resolution. Extensive experiments on various vision-language tasks validate the effectiveness of our method.
\end{abstract}  

\section{Introduction}
\label{sec:intro}
Visual Large Language Models (VLLMs) represent a powerful class of models capable of handling vision-language tasks~\citep{Yin2023ASO,liu2023improvedllava,liu2024llavanext,Alayrac2022FlamingoAV}. There is a growing body of research focused on the application of VLLMs in real-world scenarios, where different tasks necessitate varying levels of perceptual granularity. For instance, autonomous driving systems require high resolution to capture multiple objects and intricate details~\citep{Zhou2023VisionLM,Ding2023HiLMDTH}, whereas image classification tasks involving singular, simple objects can be effectively performed at lower resolutions~\citep{Li2024FlexAttentionFE,Li2023MonkeyIR,Zhang2024InternLMXComposer25AV}. Despite this, most existing VLLMs, \textit{e.g.}, LLaVA, pre-assume a fixed resolution for downstream tasks, which leads to sub-optimal performance~\citep{liu2023llava,liu2023improvedllava,blip2}. A direct ``\textit{exhaustive training}" strategy to adapt current VLLMs for diverse vision-language applications by training the models at different resolutions during the pre-training phase to create a series of checkpoints corresponding to various image input resolutions, followed by the selection of the most effective checkpoint for downstream tasks. While this method is viable, it incurs significant training costs. Consequently, we pose the first research question (\textit{\hypertarget{RQ1}{\textbf{RQ1}}}):

\textit{For a given vision-language task, how can we accurately determine the optimal resolution \textbf{without such exhaustive training} for VLLMs?}

\begin{figure}
    \centering     
    \includegraphics[width=0.96\linewidth]{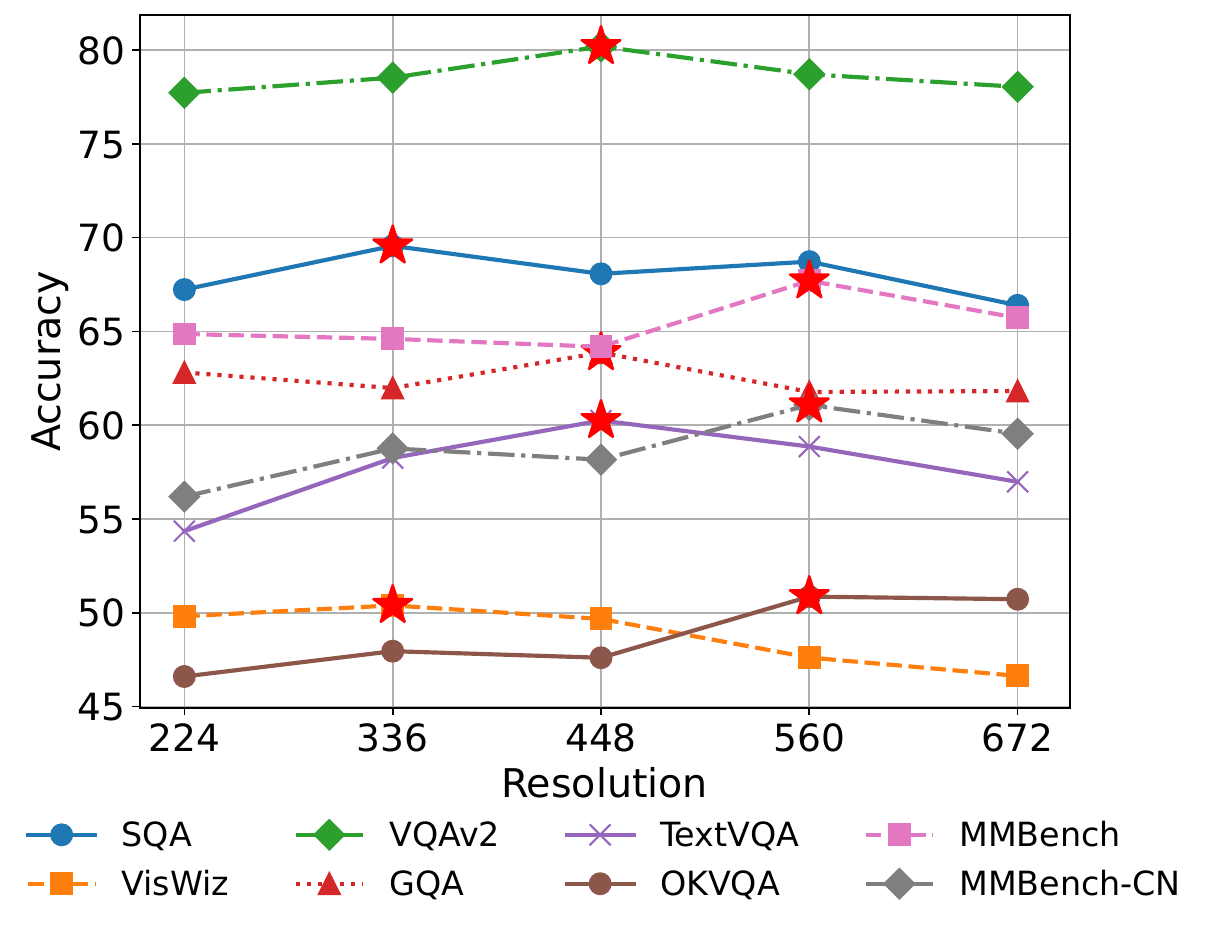}
    \caption{Resolution preference across eight tasks; \textcolor{red}{\ding{72}} marks the optimal resolution for each task.}
\label{fig:intro-fig}
\end{figure}

To answer \hyperlink{RQ1}{\textbf{\textit{RQ1}}}, we conduct a comprehensive and pioneering investigation into the resolution preferences across eight widely-studied vision-language tasks, utilizing VLLMs with five varying input image resolutions, as shown in Figure~\ref{fig:intro-fig}. Our findings reveal that directly choosing the lowest ($224^2$) and highest ($672^2$) resolution leads to subpar performance across tasks. On the other hand, we observe diverse preferences for the intermediate resolutions, with optimal choices scattered among $336^2$, $448^2$, and $560^2$.

To determine the resolution preference for different tasks, we propose two heuristic methods: \ding{182}~image complexity, which measures the intrinsic complexity of a given image [\ding{118}~Section~\ref{subsubsec:method-image-complexity}]. \ding{183}~uncertainty variance, which measures the variance of uncertainty in the model predictions at different image input resolutions [\ding{118}~Section~\ref{subsubsec:measuring-uncertainty-variance}].
Through empirical analysis across eight vision-language tasks, we find that both the complexity scores and model uncertainty variance exhibit a generally positive correlation with the preferred resolution for each task. Building on this insight, we propose an empirical formula integrating both heuristics to determine the optimal resolution for each vision-language task [\ding{118}~Section~\ref{subsubsec:method-empirical-formula}].
We utilize three reference tasks to optimize a single hyperparameter of this empirical formula, and the fitting results across five additional tasks affirm its generalizability.

Once the optimal resolution for a given vision-language task is identified, the next step is adapting the current VLLM to the identified resolution. While the training-free method exists for resolution extension, we empirically find it would lead to performance degradation, suggesting that training-based approaches are essential. However, re-training a VLLM with another resolution from scratch incurs significant costs. This prompts our second research question (\textit{\hypertarget{RQ2}{\textbf{RQ2}}}):

\textit{How can we \textbf{efficiently} adapt a pre-trained VLLM to the designated resolution without compromising performance?}

To tackle this problem, we propose a post-training strategy that extends the image input resolution of an existing VLLM checkpoint. We conduct a preliminary experiment to identify which parameters within the VLLM are crucial for performance enhancement. Based on the findings, we propose a parameter-efficient fine-tuning (PEFT) approach, which only requires updating a few parameters in each VLLM component: the positional embedding parameters of the visual encoder, the projector parameters, and the LoRA adapter parameters of the LLM backbone. Empirical studies demonstrate that our method achieves a compelling efficiency-performance trade-off.
In summary, this paper has the following contributions:
\begin{itemize}
[leftmargin=*,itemsep=1pt]
    \item \textbf{Novel Discovery.} Through a comprehensive and pioneering investigation, we discover that different vision-language tasks prefer distinct resolutions.
    \item \textbf{Empirical Formula.} We find these preferences correlated with image complexity and model uncertainty variance on samples at different input image resolutions. We then propose an empirical formula to adaptively determine the optimal resolution for various downstream vision-language tasks without exhaustively training VLLMs.
    \item \textbf{Efficient Adaptation.} We introduce a PEFT approach to extend the input image resolution of LLaVA through post-training, containing three components, including vision module PEFT, language module PEFT, and the projector tuning.
\end{itemize}
\section{Related Work}



\paragraph{VLLM Architectures.}
Vision Large Language Models, as one of the most capable and popular solutions to multimodal tasks, extend the reasoning and generating ability of Large Language Models (LLMs) beyond language modalities to encompass inputs such as images, video, and audio~\citep{mckinzie2024mm1, tong2024cambrian, xue2024xgen}. 
VLLMs can be categorized according to their architecture~\citep{liu2023llava,Driess2023PaLMEAE,fuyu,Team2024ChameleonME}. The encoder-decoder VLLM paradigm, which is the focus of this study, introduces additional multimodal encoders (typically a vision encoder like ViT) and a modality connector to project multimodal features into the spaces interpretable by language models. The implementations of the modality connector vary; common approaches include a projector that directly maps visual features to the language model's embedding space~\citep{liu2024llavanext,liu2023improvedllava,liu2023llava}, or a resampler that compresses visual features, possibly using cross-gated attention layers, before integrating them into the LLM decoder~\citep{Alayrac2022FlamingoAV,Awadalla2023OpenFlamingoAO,Li2023OtterAM}. Our work primarily considers LLaVA-style VLLMs, which adopt an encoder-decoder architecture with a projector connector.

\paragraph{Resolution Sensitivity in Visual Models.}
The sensitivity of visual models to input image resolution is a well-established phenomenon. Convolutional Neural Networks (CNNs) inherently leverage inductive biases like local receptive fields and hierarchical feature extraction, tying their performance to spatial information density, where higher resolutions often improve accuracy~\citep{raghu2021vision, borji2021enhancing, sabottke2020effect}. Techniques like dilated convolutions were developed to manage varying receptive field sizes~\citep{chen2017rethinking}. Vision Transformers (ViTs), processing images as sequences of patches, also exhibit distinct resolution sensitivities influenced by patch size and pre-training configurations, often struggling with resolutions unseen during training~\citep{fan2024vitar,dehghani2023patch}. Adapting positional embeddings is a common strategy to mitigate this for ViTs~\citep{qwen_vl, blip2, tian2023resformer}. While VLLMs inherit this sensitivity, the interaction with language understanding in multimodal tasks introduces new complexities. Our work aims to quantify and address this specific challenge by proposing a heuristic-driven optimization framework for VLLMs.

\begin{table*}[hbtp]
\centering
\caption{Key Distinctions: Our Task-Aware Adaptation vs. Native Dynamic Resolution VLLMs}
\label{tab:sota_comparison_main}
\resizebox{\linewidth}{!}{
    \begin{tabular}{l|l|l}
    \toprule
    \textbf{Comparative Aspect} & \textbf{Our Method (Task-Aware Adaptation)} & \textbf{Native Dynamic Resolution VLLMs} \\
    \midrule
    \textbf{Resolution Handling} & 
        \ding{51} Task-Optimized (Post-hoc PEFT) & 
        \ding{51} Inherent (Architectural Design) \\
    \midrule
    \textbf{Optimal Resolution for Task} & 
        \ding{51} Explicitly Selected (Heuristic-driven) & 
        \ding{55} Generally Implicit / Not Primary Focus \\
    \midrule
    \textbf{Adaptation Approach} & 
        \ding{51} Lightweight PEFT (on existing models) & 
        \ding{55} Extensive Pre-training / Full Fine-Tuning \\
    \midrule
    \textbf{Base Model Architecture} & 
        \ding{55} Unchanged (Adapts standard VLLMs) & 
        \ding{51} Often Modified (e.g., RoPE, specialized ViTs) \\
    \midrule
    \textbf{Resolution Decision Informed by Textual Context} & 
        \ding{51} via model uncertainty with text & 
        \ding{55} Typically visual input properties only \\
    \midrule
    \textbf{Adaptation Cost} & 
        \ding{51} Low (Efficient for existing checkpoints) & 
        \ding{55} High (Resource-intensive initial training) \\
    \bottomrule
    \end{tabular}
}
\end{table*}

\begin{figure*}
    \centering     
    \includegraphics[width=\linewidth]{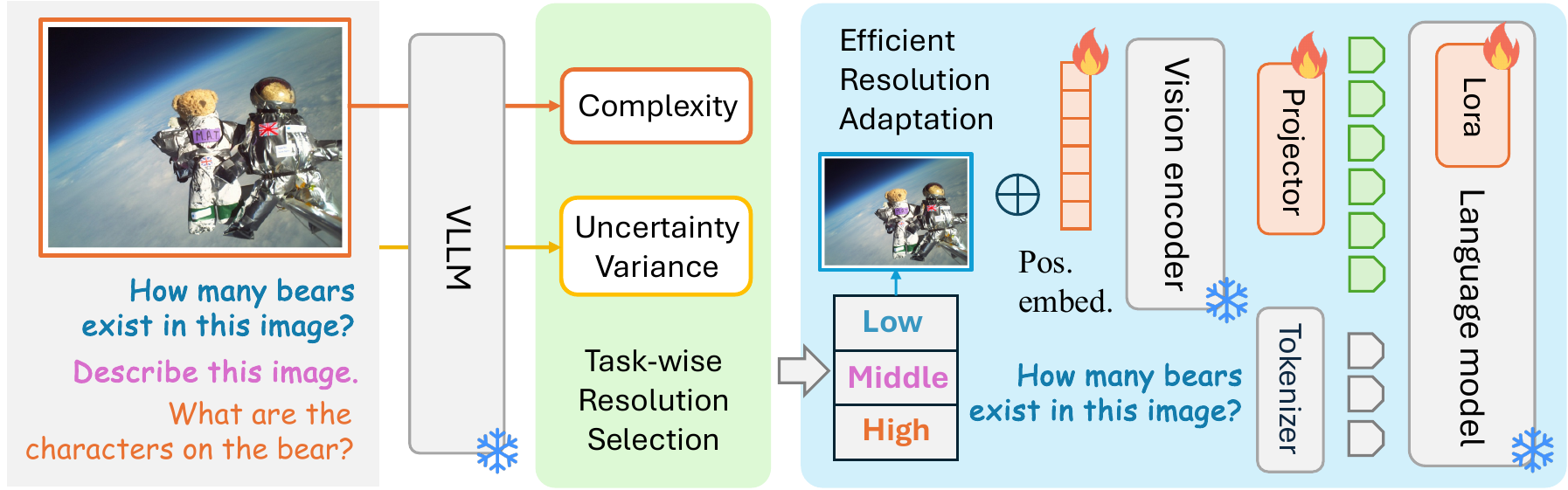}
    \caption{{Our method comprises two components: the first component identifies the optimal image input resolution for a given vision-language task (depicted in \textcolor{green}{green}), while the second component adapts the VLLM to the selected image input resolution (depicted in \textcolor{blue}{blue}).}}
    \label{fig:method-framework}
\end{figure*}

\paragraph{Strategies for Adapting VLLMs to Varying Resolutions.}
To address fixed-resolution limitations, various strategies exist. Many recent VLLMs natively support dynamic resolutions via architectural innovations (e.g., 2D RoPE in Qwen2VL~\citep{wang2024qwen2vl}, efficient high-resolution processing in MiniCPM~\citep{yao2024minicpm_v}, or varied aspect ratio handling in LLaVA-UHD~\citep{guo2025llava_uhd}), but these typically require extensive pre-training. Other techniques focus on processing high-resolution inputs through methods like image patching~\citep{chen2024expanding, Dong2024InternLMXComposer24KHDAP}, region-aware mechanisms~\citep{Wu2023VGV,Zhao2024MGLLaVATM, Zhang2023TowardsPS}, or by optimizing computational costs~\citep{Li2024FlexAttentionFE}.

Our approach differs significantly by enabling lightweight, post-training adaptation of \textit{existing} VLLM checkpoints. We first determine an optimal task-level resolution using interpretable heuristics and then efficiently adapt the model using a PEFT strategy, without architectural changes or retraining from scratch. This offers a practical pathway to enhance existing models. Key differences between our method and native dynamic resolution VLLMs are summarized in Table~\ref{tab:sota_comparison_main}. Further details on these dynamic resolution models are in Appendix~\ref{appendix:sec:dynamic_high_res_vllms}.

\section{Methodology}
\label{sec:method}

This section elaborates on our proposed methodology.  Section~\ref{subsec:method-framework} presents an overview, followed by a detailed explanation of each component in Sections~\ref{subsec:method-task-wise-selection} and~\ref{subsec:method-PEFT-adaptation}.

\subsection{Method Framework}
\label{subsec:method-framework}

Figure~\ref{fig:method-framework} illustrates our proposed two-stage approach.
The first stage, task-specific resolution selection, aims to identify the optimal input resolution for a given vision-language task. This is achieved by first employing two heuristic metrics: image complexity (detailed in Section~\ref{subsubsec:method-image-complexity}) and model uncertainty variance across different resolutions (Section~\ref{subsubsec:measuring-uncertainty-variance}). Building on these heuristics, we then introduce an empirical formula (Section~\ref{subsubsec:method-empirical-formula}) to determine this optimal task-level resolution.
Once the optimal resolution is identified for a particular task, the second stage, VLLM adaptation, adjusts the pre-trained VLLM to operate effectively at this new resolution. This adaptation is performed using a PEFT strategy (detailed in Section~\ref{subsec:method-PEFT-adaptation}), which involves post-training an existing VLLM checkpoint without requiring a full retraining from scratch. Subsequently, this adapted model is deployed to process all samples, including previously unseen ones, for that specific task at the determined optimal resolution.

\subsection{Task-wise Optimal Resolution Selection}
\label{subsec:method-task-wise-selection}

As highlighted in Section~\ref{sec:intro}, different vision-language tasks have varying requirements for the perceptual capacity of VLLMs. Therefore, it is critical to do task-wise resolution selection. While tuning VLLMs at different image input resolutions and obtaining the best-performing one is feasible, it imposes heavy training costs, which leads to \hyperlink{RQ1}{\textbf{\textit{RQ1}}}. In this section, we propose a training-free method for determining the optimal resolution for a specific vision-language task, utilizing two heuristic approaches. 
We then derive an empirical formula to guide the resolution selection process.

\subsubsection{Measuring Image Complexity}
\label{subsubsec:method-image-complexity}
The initial stage of VLLM processing involves visual perception. Intuitively, more complex images, requiring finer perceptual granularity, may benefit from higher input resolutions. Consequently, for a given vision-language task, the inherent complexity of its associated images can serve as an indicator of resolution preference.

To quantitatively assess image complexity, we adopt the method by \citet{image_complexity}, which leverages the Minimum Description Length (MDL) principle for hierarchical pixel clustering to identify perceptually meaningful structures. Key steps involve initial MDL-based pixel clustering, followed by constructing and recursively clustering patch signatures to capture multi-scale complexity, with the final score derived from summed entropies. For a comprehensive algorithmic description, we refer the reader to the original publication~\citep{image_complexity} and their publicly available implementation\footnote{https://github.com/Lou1sM/meaningful\_image\_complexity}.

In our framework, this score, averaged across all sampled images for a given task $T$, is denoted as $C(T)$ and serves as a key heuristic (Section~\ref{subsubsec:method-empirical-formula}). We chose this recent method for its efficacy in capturing perceptual complexity and its favorable comparisons to alternatives~\citep{khan2022leveraging,machado2015computerized,redies2012phog,de2013multi}, as demonstrated in~\citet{image_complexity}.

\subsubsection{Measuring Uncertainty Variance Across Resolutions}
\label{subsubsec:measuring-uncertainty-variance}
Beyond static image complexity (Section~\ref{subsubsec:method-image-complexity}), VLLM prediction uncertainty offers insights into visual-linguistic interplay and sensitivity to resolution variations. We thus introduce a second heuristic based on model uncertainty variance.

Specifically, consider a VLLM pre-trained at a fixed resolution (e.g., $336^2$ for LLaVA). We first extend its visual encoder's capacity to handle a different, typically higher, resolution by interpolating its positional embeddings, a technique employed in prior works~\citep{qwen_vl, blip2}. Let $M_1$ denote the original model operating at its native resolution, and $M_2$ denote the same model adapted to operate at the extended resolution (without further fine-tuning at this stage). To assess uncertainty robustness, we apply random augmentations to the input images of a given task $T$ using the RandAugment algorithm~\citep{cubuk2020randaugment}. Inference is then performed on these augmented task samples using both $M_1$ and $M_2$, from which we extract the softmax probability distributions for each generated token.

Token uncertainty is quantified by information entropy: $H(p) = -\sum_{i=1}^{n} p_i \log p_i$, where $p_i$ is the $i^{th}$ token's softmax probability. Sample-level uncertainty is the average entropy of all generated tokens in an output sequence (computed independently for $M_1, M_2$). Task-level average uncertainties, $U_1(T)$ and $U_2(T)$, are then derived by averaging these sample-level uncertainties across all selected samples for task $T$. The uncertainty variance, $V(T)$, for task $T$ is the relative change:  $V(T) = \frac{U_2(T) - U_1(T)}{U_1(T)}.$
A higher $V(T)$ indicates greater sensitivity of model uncertainty to resolution changes for task $T$. This $V(T)$ is the second heuristic for our empirical formula (Section~\ref{subsubsec:method-empirical-formula}).

This uncertainty-based heuristic offers two main advantages to complement the static image complexity: (1) by computing entropy from tokens generated by the VLLM, it inherently accounts for both visual and linguistic features during inference; and (2) it directly quantifies the variance induced by resolution changes, thereby capturing the dynamic effects of such shifts. Notably, calculating this heuristic involves extending VLLM input resolution without parameter tuning, avoiding extra training costs at this stage.

\subsubsection{Empirical Formula for Optimal Resolution Estimation}
\label{subsubsec:method-empirical-formula}
Inspired by the intuition that tasks with more complex imagery or higher resolution sensitivity (in terms of model prediction uncertainty) might benefit from increased input resolutions, we propose an empirical formula to estimate the optimal resolution for a given vision-language task. This intuition, regarding the positive correlation of image complexity and uncertainty variance with preferred resolution, is further explored and validated in Section~\ref{subsec:investigation}. The proposed formula is:
\begin{equation}
    Reso(T) = Reso_0 \cdot (1 + k \cdot C(T) \cdot V(T))
    \label{eq:empirical-formula}
\end{equation}
Here, $Reso_0$ is the VLLM's baseline input resolution (e.g., $336$ for LLaVA), serving as a reference for scaling. $C(T)$ is the average normalized image complexity for task $T$ (Section~\ref{subsubsec:method-image-complexity}), and $V(T)$ is its average uncertainty variance. The term $k$ is a user-specified, non-negative hyperparameter modulating the heuristics' combined influence. The expression $(1 + k \cdot C(T) \cdot V(T))$ thus acts as a scaling factor, adjusting $Reso_0$ based on task characteristics. The value of $k$ is determined empirically using reference tasks, as discussed in Section~\ref{subsubsec:applying_empirical_formula}.

\subsection{Parameter-efficient Resolution Adaptation}
\label{subsec:method-PEFT-adaptation}

After determining the optimal resolution for a given task, the next step is adapting the VLLM to the selected resolution. To answer \hyperlink{RQ2}{\textbf{\textit{RQ2}}}, We propose a parameter-efficient fine-tuning (PEFT) approach that performs post-training on an existing VLLM checkpoint, thus avoiding retraining from scratch.

As depicted in Figure~\ref{fig:method-framework}, existing VLLMs (e.g., LLaVA) consist of three main components: a visual encoder,  a projector mapping visual features to the text embedding space, and an LLM backbone generating language tokens. Increasing input resolution introduces more image patches, causing incompatibility with the original position embeddings.
To address this, we interpolate the position embeddings from the initial number of patches (e.g., $24^2$) to the extended number (e.g., $32^2$), following previous research~\citep{qwen_vl, blip2}. Although this allows the VLLM to process extended resolutions, performance degrades without further adaptation (as discussed in Section~\ref{subsec:method-task-wise-selection}). To counter this performance decline, we employ a PEFT method that fine-tunes three key components: (1) position embeddings within the visual encoder, essential for handling additional patches; (2) the lightweight projector parameters; and (3)  the parameters of the LoRA adapters integrated into the LLM backbone. By keeping all other parameters frozen, the PEFT approach offers an efficient method for adaptation. Figure~\ref{fig:method-framework} provides a visual representation of the components that are fine-tuned versus those that remain frozen.

\begin{table*}
\centering
\caption{A comprehensive investigation conducted to explore resolution preferences across eight vision-language tasks. For each task, the accuracy scores corresponding to five different resolutions are presented.}
\resizebox{\linewidth}{!}{
\begin{tabular}{l|cccccccc}
\toprule
Resolution & SciQA-IMG      & VizWiz         & VQAv2          & GQA            & TextVQA        & OKVQA          & MMBench        & MMBench-CN\\
\midrule
$224\times 224$        & $67.23$          & $49.81$          & $77.72$          & $62.81$          & $54.35$          & $46.60$          & $64.86$          & $56.19$          \\
$336\times 336$         & $\mathbf{69.56}$ & $\mathbf{50.39}$ & $78.53$          & $61.98$          & $58.25$          & $47.95$          & $64.60$          & $58.76$         \\
$448\times 448$         & $68.07$          & $49.67$          & $\mathbf{80.19}$ & $\mathbf{63.87}$ & $\mathbf{60.25}$ & $47.60$          & $64.18$          & $58.16$    \\
$560\times 560$         & $68.72$          & $47.61$          & $78.71$          & $61.77$          & $58.86$          & $\mathbf{50.86}$ & $\mathbf{67.70}$ & $\mathbf{61.08}$ \\
$672\times 672$        & $66.39$          & $46.63$          & $78.04$          & $61.82$          & $56.98$          & $50.72$          & $65.72$          & $59.54$      \\
\bottomrule
\end{tabular}
}
\label{tab:results-investigation}
\end{table*}

\begin{table*}
\centering
\caption{Distributions of image complexity and uncertainty variance across eight tasks.}
\resizebox{\linewidth}{!}{\begin{tabular}{l|cc|ccc|ccc} 
\toprule
                                           & vizwiz & SciQA-IMG          & TextVQA & GQA    & VQAv2    & OKVQA  & MMBench & MMBench-CN  \\ 
\midrule
Resolution Preference                     & \multicolumn{2}{c|}{$336 \times 336$}   & \multicolumn{3}{c|}{$448 \times 448$}    & \multicolumn{3}{c}{$560 \times 560$}        \\
\midrule
Complexity (C)                            & $0.2191$ & $0.1437$             & $0.2919$  & $0.3236$ & $0.3017$   & $0.3112$ & $0.2323$  & $0.2329$      \\
Average        & \multicolumn{2}{c|}{$0.1814$} & \multicolumn{3}{c|}{$0.3058$} & \multicolumn{3}{c}{$0.2588$}     \\
\midrule
Uncertainty Variance (V)                  & $1.83$\% & $6.47$\%             & $4.88$\%  & $5.34$\% & $5.26$\%   & $6.72$\% & $10.79$\% & $10.45$\%     \\
Average & \multicolumn{2}{c|}{$4.15$\%} & \multicolumn{3}{c|}{$5.16$\%} & \multicolumn{3}{c}{$9.32$\%}     \\
\midrule
C $\times$ V           & $0.0040$ & $0.0093$             & $0.0142$  & $0.0173$ & $0.0159$   & $0.0209$ & $0.0251$  & $0.0243$      \\
Average               & \multicolumn{2}{c|}{$0.0067$} & \multicolumn{3}{c|}{$0.0158$} & \multicolumn{3}{c}{$0.0234$}     \\
\bottomrule
\end{tabular}}
\label{table:results-complexity-uncertainty-variance-distribution}
\end{table*}

\section{Experiments}
This section presents the empirical evaluation of our proposed method. We first introduce the implementation details in Section~\ref{subsec:implementation-details}, followed by an in-depth analysis of the results, including the investigation into resolution preferences, task-wise resolution selection, and the findings from the ablation study in Section~\ref{subsec:investigation}, \ref{subsec:task-wise-resolution-selection}, and~\ref{subsec:ablation-study}, respectively.

\subsection{Implementation Details}
\label{subsec:implementation-details}
\textbf{VLLM Selection.} For our experiments, we select the LLaVA-1.5-7B checkpoint~\cite{liu2023llava} as the representative VLLM for evaluation.

\noindent\textbf{Resolution Configurations.}
We explore five image resolutions: $224^2$, $336^2$, $448^2$, $560^2$, and $672^2$. These values cover the resolution spectrum commonly used in previous studies~\cite{liu2023llava, liu2023improvedllava}.

\noindent\textbf{Vision-Language Tasks.} Our evaluation encompasses eight vision-language tasks, with details introduced in Appendix~\ref{subsec:appendix-vision-language-tasks}.

\noindent\textbf{Baseline Methods.} In addition to the original LLaVA model, we compare our method with several state-of-the-art approaches. Besides, we report the performance of position embedding interpolation as a representative of the training-free methods to extend the image input resolution of VLLMs. The details are introduced in Appendix~\ref{subsec:appendix-baseline-methods}.

\noindent\textbf{Post-training Details.}
To initialize the position embedding parameters of the visual encoder (Vision Transformer) in LLaVA during resolution adaptation, we employ extended position embeddings derived through positional embedding interpolation, as described in Appendix~\ref{subsec:appendix-baseline-methods}. Following the instructions provided by the LLaVA authors\footnote{\url{https://github.com/haotian-liu/LLaVA/tree/main?tab=readme-ov-file\#train}}, we concentrate on stage 2 fine-tuning, incorporating the additional parameters for position embeddings in the visual encoder, alongside the LoRA adapter and projector parameters. The fine-tuning process utilizes images from five datasets: COCO~\cite{coco}, GQA~\cite{hudson2019gqa}, OCR-VQA~\cite{mishra2019ocrvqa}, TextVQA~\cite{textvqa}, and Visual Genome~\cite{krishna2017VisualGenome}. For more details on the construction of the image-text pairs used in training, we refer readers to \cite{liu2023improvedllava}. It is crucial to note that this post-training stage is designed solely to adapt the VLLM to the newly selected input resolution, not to specialize it for a particular task.

Further details regarding the overall method implementation and our PEFT setup are provided in Appendix~\ref{subsec:appendix-method} and~\ref{subsec:appendix-PEFT}, respectively.

\subsection{Analyzing Resolution Preferences Across Vision-Language Tasks}
\label{subsec:investigation}

We systematically analyze resolution preferences across vision-language tasks (Table~\ref{tab:results-investigation}), revealing two key findings: \ding{182} Performance is suboptimal at very low ($224^2$) or very high ($672^2$) resolutions—low resolution limits visual detail capture, while high resolution disrupts adaptation and introduces irrelevant tokens. \ding{183} Optimal resolutions lie in the mid-range ($336^2$, $448^2$, $560^2$), varying by task, which underscores the need for task-specific selection.

\begin{figure}[h!]
    \centering
    \includegraphics[width=0.95\linewidth]{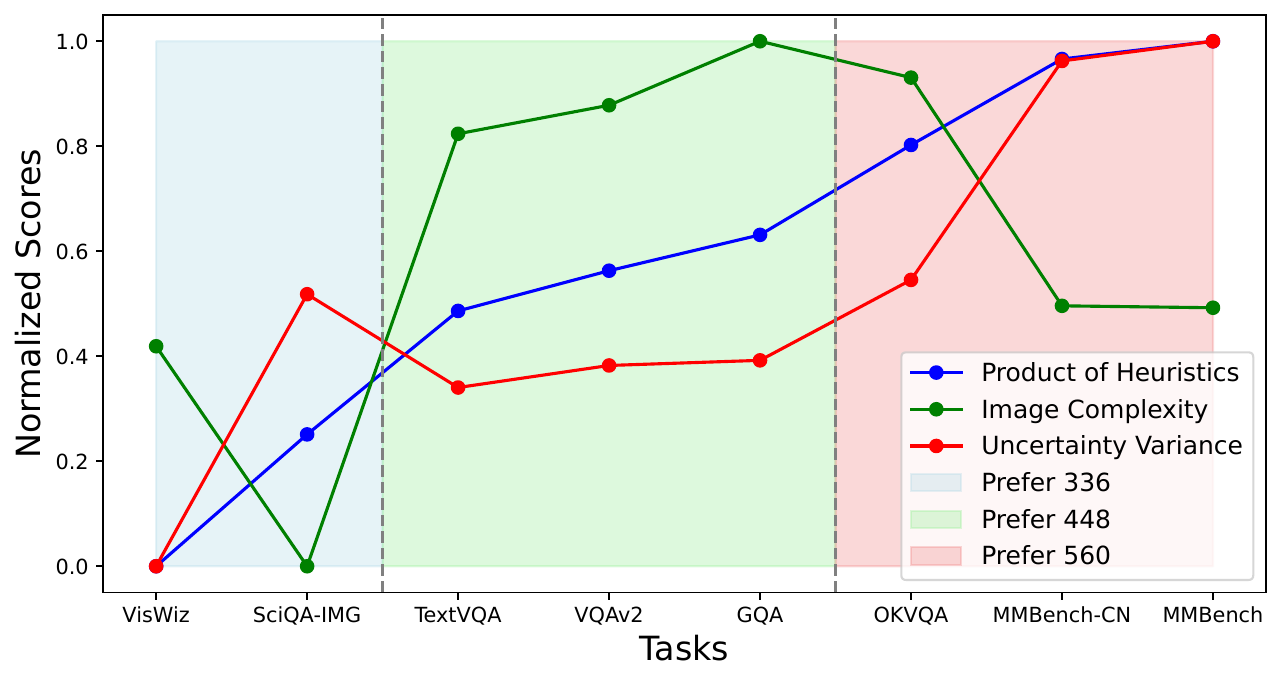}
    \caption{Correlation of heuristic metrics with preferred task resolution. The product of $C(T)$ and $V(T)$ exhibits a more consistent correlation compared to individual heuristics. All metrics are normalized for visualization.}
    \label{fig:heuristic_correlation_resolution}
\end{figure}


\begin{figure*}[!htbp]
    \centering
    \begin{subfigure}[b]{0.3\textwidth}
        \includegraphics[width=0.81\textwidth]{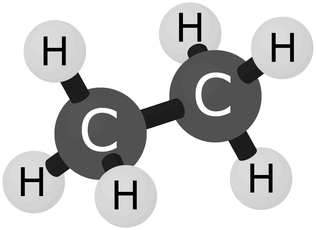}
        \caption{Single and simple object: Ethane is (). A. an elementary substance B. a compound}
        \label{fig:imageA}
    \end{subfigure}
    \hfill
    \begin{subfigure}[b]{0.3\textwidth}
        \includegraphics[width=\textwidth]{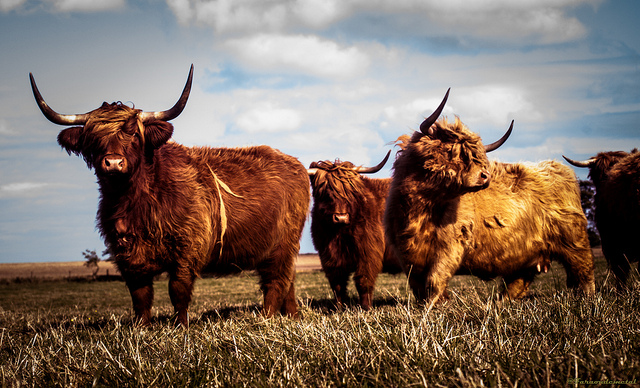}
        \caption{Middle-level complexity: Are all the animals the same?}
        \label{fig:imageB}
    \end{subfigure}
    \hfill
    \begin{subfigure}[b]{0.3\textwidth}
        \includegraphics[width=0.92\textwidth]{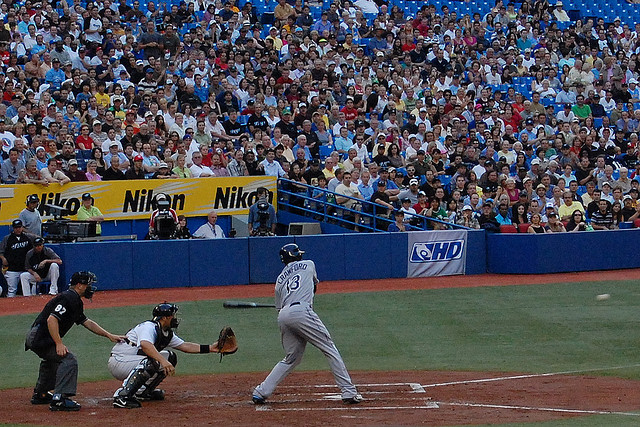}
        \caption{Multiple objects: What is the brand being advertised?}
        \label{fig:imageC}
    \end{subfigure}
    \caption{We select three reference tasks with images in different levels of complexity to optimize the hyperparameter in Equation~\ref{eq:empirical-formula}.}
    \label{fig:three-reference-tasks}
\end{figure*}

\begin{figure*}[htbp]
    \centering
    \begin{subfigure}[b]{0.45\textwidth}
        \includegraphics[width=\textwidth]{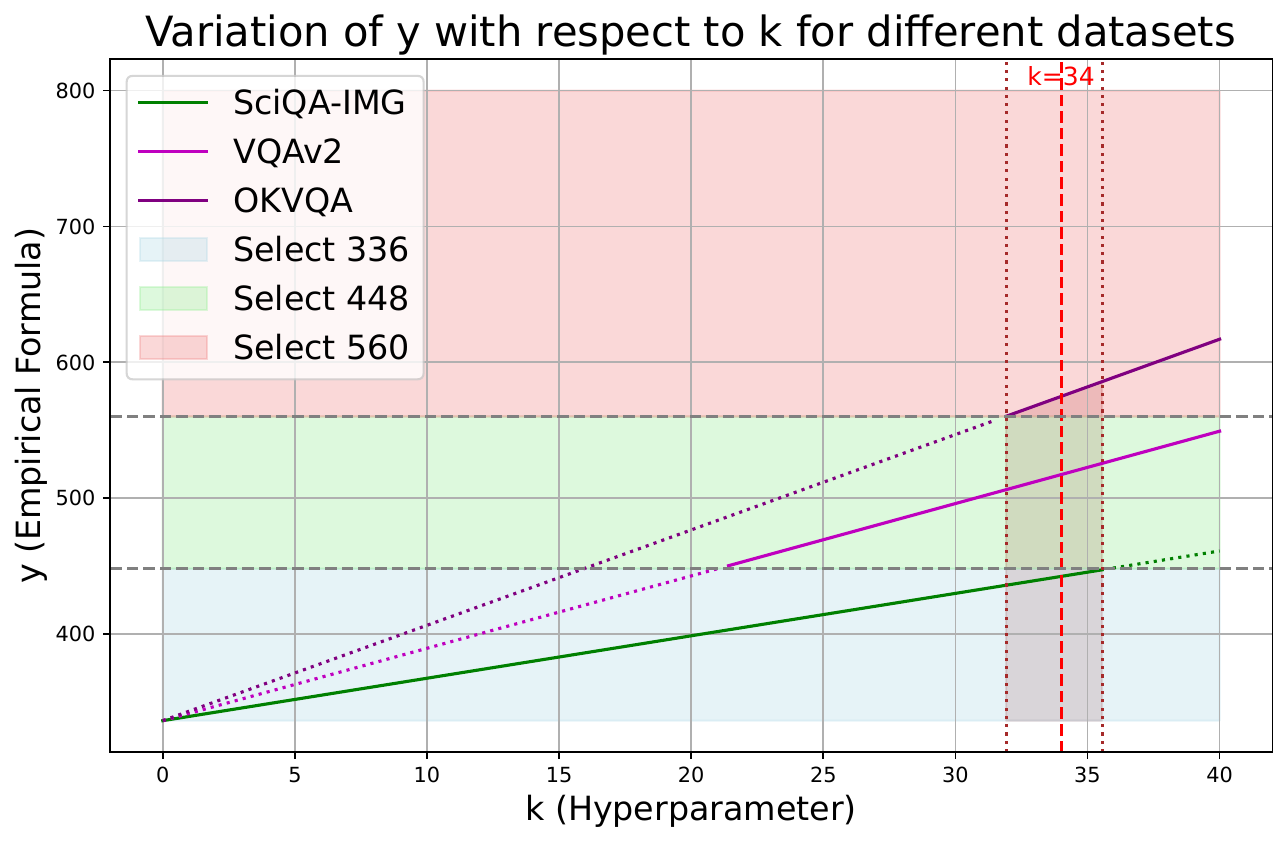}
        \caption{Optimization of the hyperparameters in the empirical formula using three reference tasks.}
        \label{fig:empirical-formula-hyperparameter-tuning}
    \end{subfigure}
    \hfill
    \begin{subfigure}[b]{0.45\textwidth}
        \includegraphics[width=\textwidth]{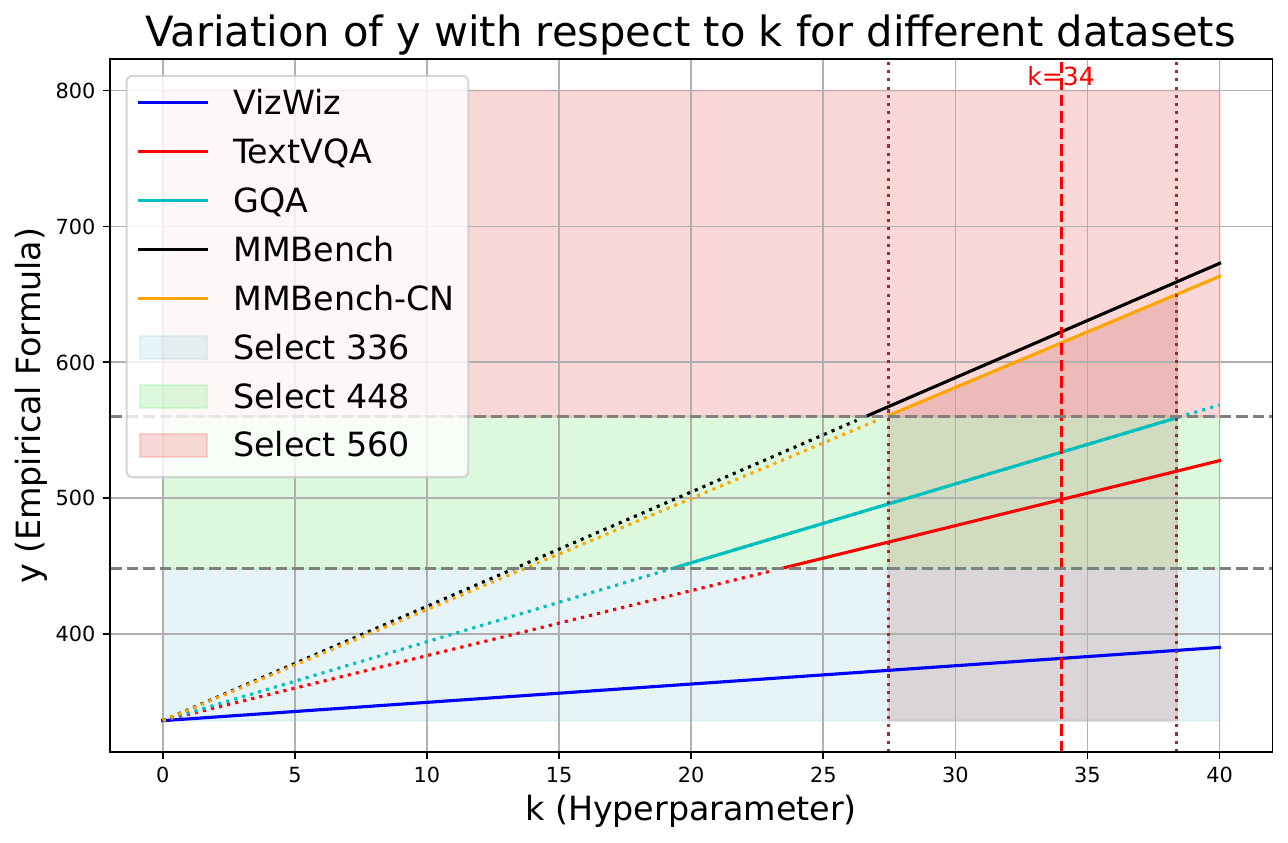}
        \caption{The empirical formula demonstrates effective generalization across five vision-language tasks.}
        \label{fig:empirial-formula-generalize}
    \end{subfigure}
    \caption{Applying the empirical formula to determine the optimal resolution for vision-language tasks.}
    \label{fig:emprical-formula}
\end{figure*}

After identifying task-specific resolution preferences, we explore the correlation between optimal resolutions and our proposed heuristics of image complexity and uncertainty variance, as shown in Table~\ref{table:results-complexity-uncertainty-variance-distribution}. We can draw the following conclusions: \ding{182} No increasing trend is observed between $448^2$ and $560^2$ in image complexity, but a noticeable gap exists between $336^2$ and $448^2$, suggesting that image complexity differentiates tasks favoring $336^2$ from those preferring higher resolutions. \ding{183} There is a positive correlation between preferred resolution and uncertainty variance across tasks, with an upward trend showing that uncertainty variance reliably indicates resolution preference. \ding{184} Some exceptions exist, e.g., GQA prefers lower resolution than MMbench but has higher image complexity, and SciQA-IMG has higher uncertainty variance but favors a lower resolution than TextVQA. Multiplying the scores of two heuristics provides a more consistent correlation, as shown in Figure~\ref{fig:heuristic_correlation_resolution}.

\subsection{Evaluating Heuristic-Based Task-Specific Resolution Selection}
\label{subsec:task-wise-resolution-selection}

The investigation presents the correlation between task-specific resolution preferences and two heuristics. This section describes hyperparameter determination for our empirical formula and summarizes the performance of models using this strategy.

\begin{table*}[htbp]
\centering
\caption{Comparison between our method and baseline approaches, highlighting the best scores in bold. \oneS indicates that the training images or annotations of the datasets were observed during training.}
\begin{lrbox}{\myentiretablebox}
\begin{tabular}{llll|cccccc}
\toprule
Method       & LLM        & Resolution         & Post-training & VQAv2 & GQA   & TextVQA & OKVQA & MMBench & MMBench-CN  \\
\midrule
BLIP-2       & Vicuna-13B & $224\times 224$                & -             & $65.00$ & $41.00$ & $42.50$   & -     & -       & -           \\
InstructBLIP & Vicuna-7B  & $224\times 224$                & -             & -     & $49.20$ & $50.10$   & -     & $36.00$   & $23.70$       \\
InstructBLIP & Vicuna-13B & $224\times 224$                & -             & -     & $49.50$ & $50.70$   & -     & -       & -           \\
Shikra\tnote{\dag}       & Vicuna-13B & $224\times 224$                & -             & $77.40$\oneS & -     & -       & -     & $58.80$   & -           \\
IDEFICS-9B   & LLaMA-7B   & $224\times 224$                & -             & $50.90$ & $38.40$ & $25.90$   & -     & $48.20$   & $25.20$       \\
IDEFICS-80B  & LLaMA-65B  & $224\times 224$                & -             & $60.00$ & $45.20$ & $30.90$   & -     & $54.50$   & $38.10$       \\
Qwen-VL      & Qwen-7B    & $448\times 448$                & -             & $78.80$\oneS & $59.30$\oneS & $\mathbf{63.80}$\oneS   & -     & $38.20$   & $7.40$        \\
Qwen-VL-Chat & Qwen-7B    & $448\times 448$                & -             & $78.20$\oneS & $57.50$\oneS & $61.50$\oneS   & -     & $60.60$   & $56.70$       \\
\midrule
LLaVA-1.5    & Vicuna-7B  & $336\times 336$                & -             & $78.53$\oneS & $61.98$\oneS & $58.25$   & $47.95$ & $64.60$   & $58.76$       \\
LLaVA-1.5    & Vicuna-7B  & $448\times 448$                & \ding{55}            & $77.82$\oneS & $61.29$\oneS & $56.61$   & $47.38$ & $63.32$   & $57.73$       \\
LLaVA-1.5    & Vicuna-7B  & $448\times 448$                & \ding{51}           & $\mathbf{80.19}$\oneS & $\mathbf{63.87}$\oneS & $60.25$   & $47.60$ & $64.18$   & $58.16$       \\
LLaVA-1.5    & Vicuna-7B  & $560\times 560$                & \ding{51}           & $78.71$\oneS & $61.77$\oneS & $58.86$   & $\mathbf{50.86}$ & $\mathbf{67.70}$   & $\mathbf{61.08}$       \\
\rowcolor{mylightgray}
LLaVA-1.5    & Vicuna-7B  & Adaptive & \ding{51}           & $\mathbf{80.19}$\oneS & $\mathbf{63.87}$\oneS & $60.25$   & $\mathbf{50.86}$ & $\mathbf{67.70}$   & $\mathbf{61.08}$       \\
\midrule
\graytext{LLaVA-1.5}    & \graytext{Vicuna-13B} & \graytext{$336\times 336$}                & -             & \graytext{$80.00$\oneS} & \graytext{$63.30$\oneS} & \graytext{$61.30$}   & \graytext{-}     & \graytext{$67.70$}   & \graytext{$63.60$}  \\
\bottomrule
\end{tabular}
\end{lrbox}

\begin{threeparttable}
        \resizebox{\linewidth}{!}{\usebox{\myentiretablebox}}
        \scriptsize 
        \color{lightgraytext} 
        \begin{tablenotes}
            \item[\dag] Shikra, primarily a referential dialogue model, is evaluated here in a VQA instruction-following setting for broader comparison. 
        \end{tablenotes}
    \end{threeparttable}

\label{table:results-task-wise-selection}
\end{table*}

\subsubsection{Applying the empirical formula to determine the optimal resolution}
\label{subsubsec:applying_empirical_formula}

To optimize the hyperparameter in Equation~\ref{eq:empirical-formula}, we select three reference tasks representing different visual perception requirements (Figure~\ref{fig:three-reference-tasks} shows task images). Tasks with simpler images (e.g., Figure~\ref{fig:imageA}) are considered low resolution, while complex images (e.g., Figure~\ref{fig:imageC}) require higher resolutions. Intermediate tasks (e.g., Figure~\ref{fig:imageB}) represent medium resolution. SciQA-IMG, VQAv2, and OKVQA are separately chosen to reflect low, medium, and high resolution needs.

When tuning the hyperparameter $k$, we focus on $336^2$, $448^2$, and $560^2$. The constant $Reso_0$ is set to $336$ (default LLaVA resolution). The formula selects the resolution based on the value of $k$. For instance, when the empirical formula yields $Reso(T) = 500$,  
we follow the rule of mapping it to the largest supported resolution 
that does not exceed this value, namely $448^2$.
Figure~\ref{fig:empirical-formula-hyperparameter-tuning} visualizes the relationship between hyperparameter values and selected resolutions. For simplicity, we select $k = 34$, which results in optimal resolution selection for the reference tasks. Additionally, as shown in Figure~\ref{fig:empirial-formula-generalize}, this value generalizes well to other tasks, achieving the best resolution for each.

While the empirical formula demonstrates good generalization with a fixed $k$ value, its practical application to a new task involves sampling a subset of data from that task to compute $C(T)$ and $V(T)$. Appendix~\ref{appendix:sec-stat-impact-on-formula} analyzes the formula's robustness to varying sample sizes, including the relationship between sampling ratio and prediction success, and the influence of heuristic distributions, offering guidance for data-limited applications.

\subsubsection{Overall results of Task-wise Adaptive Model and Baselines}

\begin{table*}[h]
\centering
\caption{Ablation Analysis of PEFT Components, \ding{55} and \ding{51} indicate whether the module is post-trained.}
\resizebox{0.8\linewidth}{!}{\begin{tabular}{llll|lll}
\toprule
Resolution & ViT PE & Projector & LoRA Adapter & VQAv2 & GQA & TextVQA  \\
\midrule
$336 \times 336$        & -             & -                & -                   &   $78.53\,(-2.07\%)$    &  $61.98\,(-2.96\%)$   & $58.25\,(-3.32\%)$    \\
$448 \times 448$        & \ding{55}            & \ding{55}               & \ding{55}                  &   $77.82\,(-2.96\%)$    &  $61.29\,(-4.04\%)$   & $56.61\,(-6.04\%)$    \\
$448 \times 448$        & \ding{51}           & \ding{55}               & \ding{55}                  &   $75.32\,(-6.07\%)$    &  $59.98\,(-6.09\%)$   & $53.44\,(-11.30\%)$    \\
$448 \times 448$        & \ding{55}            & \ding{51}              & \ding{55}                  &   $72.94\,(-9.04\%)$    &  $55.31\,(-13.40\%)$   & $51.41\,(-14.67\%)$    \\
$448 \times 448$        & \ding{55}            & \ding{51}              & \ding{51}                 &   $79.47\,(-0.90\%)$   &  $63.41\,(-0.72\%)$   & $58.06\,(-3.63\%)$    \\
$336 \times 336$        & \ding{51}             & \ding{51}                & \ding{51}                   &   $79.33\,(-1.07\%)$    &  $63.33\,(-0.85\%)$   & $58.19\,(-3.42\%)$    \\
$448 \times 448$        & \ding{51}           & \ding{51}              & \ding{51}                 &    $\mathbf{80.19}$   &  $\mathbf{63.87}$  & $\mathbf{60.25}$   \\
\bottomrule
\end{tabular}}
\label{tab:ablation-study}
\end{table*}

\begin{figure*}[h]
    \centering
    \begin{subfigure}[b]{0.3\textwidth}
        \includegraphics[width=\textwidth, height=0.65\textwidth]{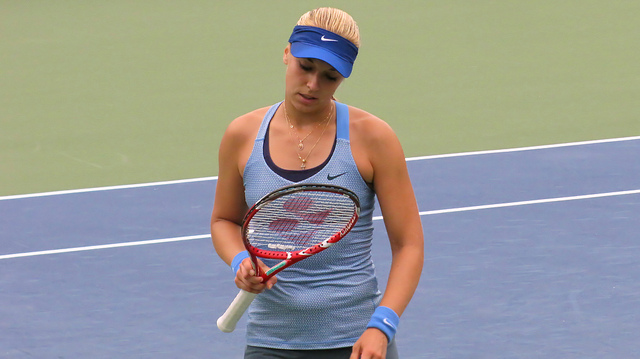}
        \caption{}
        \label{fig:caseA}
    \end{subfigure}
    \hfill
    \begin{subfigure}[b]{0.3\textwidth}
        \includegraphics[width=\textwidth, height=0.65\textwidth]{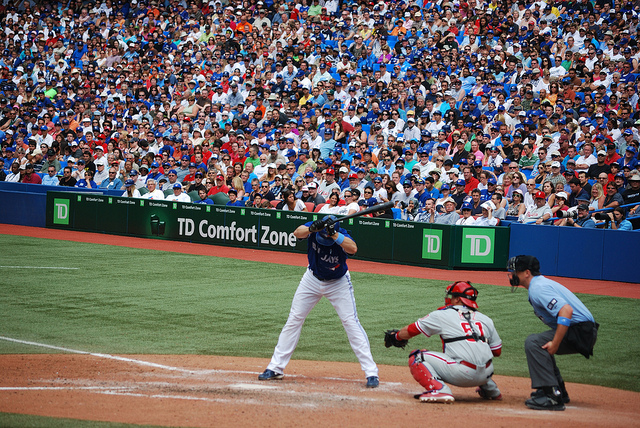}
        \caption{}
        \label{fig:caseB}
    \end{subfigure}
    \hfill
    \begin{subfigure}[b]{0.3\textwidth}
        \includegraphics[width=\textwidth, height=0.65\textwidth]{figures/gqa_case_study_uncertainty_variance.png}
        \caption{}
        \label{fig:caseC}
    \end{subfigure}
    \caption{Three case study images}
    \label{fig:case-study-figures}
\end{figure*}

Table~\ref{table:results-task-wise-selection} presents the performance of baseline methods and LLaVA variants across six tasks that demand high visual perception capacity from VLLMs. Among the LLaVA variants, the training-free method to extend the input resolution through PE interpolation shows performance degradation at varying levels. This confirms that the position embeddings in the visual encoder and LLM backbone in LLaVA cannot fully adapt to the increased number of image tokens without post-training. On the other hand, the task-wise adaptive LLaVA variant, which optimally selects the input resolution for each task, achieves the best overall performance compared to fixed-resolution LLaVA variants, regardless of whether the resolution is $336^2$, $448^2$, or $560^2$. Notably, the task-wise adaptive LLaVA variant with a 7B backbone performs comparably to the 13B variant, underscoring the importance of adaptive perception capacity in VLLMs.

When comparing the task-wise adaptive LLaVA variant with other state-of-the-art baselines, it outperforms all but the TextVQA task. In the case of TextVQA, the Qwen-VL and Qwen-VL-Chat methods have observed training images or annotations of the dataset during their training. Importantly, as previous studies~\citep{mm1} have highlighted, resolution plays a crucial role during pretraining. The Qwen-VL series are pretrained at an image resolution of $448^2$, while the LLaVA variants were fine-tuned at extended image resolutions in a post-training phase with far fewer data (665K) compared to Qwen's 1.4B pretraining and 50M fine-tuning samples. Nevertheless, the task-wise adaptive LLaVA variant achieves better overall results than the Qwen-VL series.

The superior performance of the task-wise adaptive LLaVA variant across multiple vision-language tasks demonstrates that, compared to \emph{fixed-resolution} approaches, \emph{adaptive resolution selection} is more suitable for real-world applications. So far, we have verified the effectiveness of our task-wise resolution selection strategy through the generalization of the empirical formula and the overall experimental results, answering \hyperlink{RQ1}{\textbf{RQ1}}.

\subsection{Ablation Analysis of PEFT Components for Performance}
\label{subsec:ablation-study}

To evaluate the contribution of each component in our PEFT method, we conduct an ablation study (Table~\ref{tab:ablation-study}), examining the impact of tuning three key parameters: position embeddings in the visual encoder, LoRA adapters in the LLM backbone, and projector parameters. We also assess whether performance gains stem from the additional training epoch introduced by post-training by conducting full training at the original resolution ($336^2$).

Results show that tuning each component is crucial. Tuning only position embeddings or projector parameters leads to significant drops, even compared to training-free positional embedding interpolation. While jointly tuning projector parameters and LoRA adapters improves performance, it remains suboptimal without tuning position embeddings. Additionally, post-training at $336^2$ provides only marginal gains over full training or projector + LoRA tuning at $448^2$. Notably, on TextVQA, post-training at $336^2$ offers no improvement over the original checkpoint, suggesting that gains at $448^2$ primarily stem from enhanced perceptual capabilities, not extra training.
Overall, our results highlight the importance of each component in PEFT and validate its effectiveness in addressing \hyperlink{RQ2}{\textbf{RQ2}}.

\section{Case Study}
\label{sec:case_study}

\begin{table}[hbtp]
\centering
\caption{Case studies: VLLM performance with varying image complexity and question difficulty.}
\label{tab:case_studies_optimized_columns}
\resizebox{\linewidth}{!}{
\begin{tabular}{@{}l c l l@{}} 
    \toprule
    \multicolumn{4}{@{}l@{}}{\textbf{Case 1: Varying Image Complexity} (Question: "Who is standing?")} \\ 
    \cmidrule(r){1-4}
    Image & $C(T)$ & Pred. ($336^2$) & Correct Answer \\
    \midrule
    Fig.~\ref{fig:caseA}  & $11.35$ & woman (\ding{51}) & woman \\
    Fig.~\ref{fig:caseB}  & $20.62$ & umpire (\ding{55}) & batter \\
    \midrule[\heavyrulewidth]
    \multicolumn{4}{@{}l@{}}{\textbf{Case 2: Varying Question Difficulty} (Image: Fig.~\ref{fig:caseC})} \\ 
    \cmidrule(r){1-4}
    Question  & $V(T)$ & Pred. ($336^2$) & Pred. ($448^2$) \\ 
    \midrule
    Q1: "Sheet material?"  & $0.42\%$ & plastic (\ding{51}) & plastic (\ding{51}) \\
    Q2: "Stoves near tap?"  & $16.51\%$ & NO (\ding{55})  & YES (\ding{51}) \\
    \bottomrule
\end{tabular}
}
\end{table}

Table~\ref{tab:case_studies_optimized_columns} presents two illustrative case studies demonstrating the impact of our heuristics on VLLM performance. Visual inputs are in Figure~\ref{fig:case-study-figures}.

As shown in Table~\ref{tab:case_studies_optimized_columns} (top), we present the VLLM with two images of differing complexities for the same question: "Who is standing?". At the $336^2$ resolution, the model correctly identifies the "woman" in the simpler image. However, for the more intricate image with higher complexity, it fails, incorrectly predicting "umpire" instead of "batter". This suggests that more visually complex images may necessitate higher input resolutions for accurate VLLM perception.

The second case (Table~\ref{tab:case_studies_optimized_columns}, bottom) uses a single image (Fig.~\ref{fig:caseC}) but poses two questions of differing difficulty, leading to different uncertainty variances ($V(T)$). For the easier question ("What is the sheet made of?"), the VLLM provides the correct answer ("plastic") at both $336^2$ and $448^2$ resolutions. However, for the more complex question requiring finer detail ("Are there stoves near the freezer to the right of the tap?"), the model fails at $336^2$ but succeeds at the higher $448^2$ resolution. This improved performance at higher resolution for the more uncertain (difficult) question aligns with the core intuition behind our $V(T)$ heuristic, as discussed in Section~\ref{subsubsec:method-empirical-formula}.

\section{Conclusion}
In this paper, we take a step towards adapting VLLMs to real-world applications by providing an in-depth investigation of resolution preferences in different vision-language tasks. Based on the findings, we introduce an empirical formula that combines image complexity and uncertainty variance to make task-specific resolution selection without the need for retraining. Additionally, we propose a PEFT approach, enabling extension of the image input resolution for existing VLLM checkpoints. We expect that our research will offer valuable insights for the VLLM research community.

\newpage
\section*{Limitations \& Future Work}
Our current work has several limitations. Due to computational constraints in an academic environment, we were unable to conduct experiments with larger LLM backbones or retrain models from scratch. This restricts the scope of comparison, particularly against methods requiring extensive pretraining. Moreover, our proposed approach focuses on task-level resolution selection. Future work will explore more granular resolution strategies, such as dynamic sample-level resolution adaptation, which could further improve performance for heterogeneous tasks.

\section*{Ethical Statement}

This study leverages publicly available datasets (e.g., VQAv2, GQA, TextVQA, OKVQA, MMBench) and pre-trained models (e.g., LLaVA) for evaluation and experimentation. These datasets and models are widely recognized benchmarks in the vision-language research community, distributed under licenses permitting academic and non-commercial use. All artifacts were used in accordance with their intended purposes, without modifications or new data collection. The dataset creators’ documentation ensures compliance with ethical guidelines, including the absence of personally identifiable or offensive content.

No ethics review board approval was required, as this research does not involve human subject data or sensitive information. However, we acknowledge that the underlying datasets may contain biases or inaccuracies, which could affect model fairness and generalization. Future research should explore bias mitigation strategies to ensure fair and responsible deployment of vision-language models. The derivative findings, such as task-specific resolution adaptation strategies, remain compatible with the original licenses and intended use.

\section*{Acknowledgement}

This project is partially supported by the Honda Research Institute USA.


\bibliography{main}

\begin{thebibliography}{61}
\providecommand{\natexlab}[1]{#1}

\bibitem[{fuy()}]{fuyu}

\newblock \href {https://www.adept.ai/blog/fuyu-8b/} {Fuyu-8b: A multimodal architecture for ai agents}.

\bibitem[{Alayrac et~al.(2022)Alayrac, Donahue, Luc, Miech, Barr, Hasson, Lenc, Mensch, Millican, Reynolds, Ring, Rutherford, Cabi, Han, Gong, Samangooei, Monteiro, Menick, Borgeaud, Brock, Nematzadeh, Sharifzadeh, Binkowski, Barreira, Vinyals, Zisserman, and Simonyan}]{Alayrac2022FlamingoAV}
Jean-Baptiste Alayrac, Jeff Donahue, Pauline Luc, Antoine Miech, Iain Barr, Yana Hasson, Karel Lenc, Arthur Mensch, Katie Millican, Malcolm Reynolds, Roman Ring, Eliza Rutherford, Serkan Cabi, Tengda Han, Zhitao Gong, Sina Samangooei, Marianne Monteiro, Jacob Menick, Sebastian Borgeaud, Andy Brock, Aida Nematzadeh, Sahand Sharifzadeh, Mikolaj Binkowski, Ricardo Barreira, Oriol Vinyals, Andrew Zisserman, and Karen Simonyan. 2022.
\newblock Flamingo: a visual language model for few-shot learning.
\newblock \emph{ArXiv}, abs/2204.14198.

\bibitem[{Awadalla et~al.(2023)Awadalla, Gao, Gardner, Hessel, Hanafy, Zhu, Marathe, Bitton, Gadre, Sagawa, Jitsev, Kornblith, Koh, Ilharco, Wortsman, and Schmidt}]{Awadalla2023OpenFlamingoAO}
Anas Awadalla, Irena Gao, Josh Gardner, Jack Hessel, Yusuf Hanafy, Wanrong Zhu, Kalyani Marathe, Yonatan Bitton, Samir~Yitzhak Gadre, Shiori Sagawa, Jenia Jitsev, Simon Kornblith, Pang~Wei Koh, Gabriel Ilharco, Mitchell Wortsman, and Ludwig Schmidt. 2023.
\newblock Openflamingo: An open-source framework for training large autoregressive vision-language models.
\newblock \emph{ArXiv}, abs/2308.01390.

\bibitem[{Bai et~al.(2023)Bai, Bai, Yang, Wang, Tan, Wang, Lin, Zhou, and Zhou}]{qwen_vl}
Jinze Bai, Shuai Bai, Shusheng Yang, Shijie Wang, Sinan Tan, Peng Wang, Junyang Lin, Chang Zhou, and Jingren Zhou. 2023.
\newblock Qwen-vl: A frontier large vision-language model with versatile abilities.
\newblock \emph{arXiv preprint arXiv:2308.12966}.

\bibitem[{Borji(2021)}]{borji2021enhancing}
Ali Borji. 2021.
\newblock Enhancing sensor resolution improves cnn accuracy given the same number of parameters or flops.
\newblock \emph{arXiv preprint arXiv:2103.05251}.

\bibitem[{Chen et~al.(2023)Chen, Zhang, Zeng, Zhang, Zhu, and Zhao}]{chen2023shikra}
Keqin Chen, Zhao Zhang, Weili Zeng, Richong Zhang, Feng Zhu, and Rui Zhao. 2023.
\newblock Shikra: Unleashing multimodal llm's referential dialogue magic.
\newblock \emph{arXiv preprint arXiv:2306.15195}.

\bibitem[{Chen et~al.(2017)Chen, Papandreou, Schroff, and Adam}]{chen2017rethinking}
Liang-Chieh Chen, George Papandreou, Florian Schroff, and Hartwig Adam. 2017.
\newblock Rethinking atrous convolution for semantic image segmentation.
\newblock \emph{arXiv preprint arXiv:1706.05587}.

\bibitem[{Chen et~al.(2024)Chen, Wang, Cao, Liu, Gao, Cui, Zhu, Ye, Tian, Liu et~al.}]{chen2024expanding}
Zhe Chen, Weiyun Wang, Yue Cao, Yangzhou Liu, Zhangwei Gao, Erfei Cui, Jinguo Zhu, Shenglong Ye, Hao Tian, Zhaoyang Liu, et~al. 2024.
\newblock Expanding performance boundaries of open-source multimodal models with model, data, and test-time scaling.
\newblock \emph{arXiv preprint arXiv:2412.05271}.

\bibitem[{Cubuk et~al.(2020)Cubuk, Zoph, Shlens, and Le}]{cubuk2020randaugment}
Ekin~D Cubuk, Barret Zoph, Jonathon Shlens, and Quoc~V Le. 2020.
\newblock Randaugment: Practical automated data augmentation with a reduced search space.
\newblock In \emph{Proceedings of the IEEE/CVF conference on computer vision and pattern recognition workshops}, pages 702--703.

\bibitem[{Dai et~al.(2024)Dai, Li, Li, Tiong, Zhao, Wang, Li, Fung, and Hoi}]{instructblip}
Wenliang Dai, Junnan Li, Dongxu Li, Anthony Meng~Huat Tiong, Junqi Zhao, Weisheng Wang, Boyang Li, Pascale Fung, and Steven Hoi. 2024.
\newblock Instructblip: towards general-purpose vision-language models with instruction tuning.
\newblock In \emph{Proceedings of the 37th International Conference on Neural Information Processing Systems}, NIPS '23, Red Hook, NY, USA. Curran Associates Inc.

\bibitem[{De~Siqueira et~al.(2013)De~Siqueira, Schwartz, and Pedrini}]{de2013multi}
Fernando~Roberti De~Siqueira, William~Robson Schwartz, and Helio Pedrini. 2013.
\newblock Multi-scale gray level co-occurrence matrices for texture description.
\newblock \emph{Neurocomputing}, 120:336--345.

\bibitem[{Dehghani et~al.(2023)Dehghani, Mustafa, Djolonga, Heek, Minderer, Caron, Steiner, Puigcerver, Geirhos, Alabdulmohsin et~al.}]{dehghani2023patch}
Mostafa Dehghani, Basil Mustafa, Josip Djolonga, Jonathan Heek, Matthias Minderer, Mathilde Caron, Andreas Steiner, Joan Puigcerver, Robert Geirhos, Ibrahim~M Alabdulmohsin, et~al. 2023.
\newblock Patch n’pack: Navit, a vision transformer for any aspect ratio and resolution.
\newblock \emph{Advances in Neural Information Processing Systems}, 36:2252--2274.

\bibitem[{Deng et~al.(2009)Deng, Dong, Socher, Li, Li, and Fei-Fei}]{deng2009imagenet}
Jia Deng, Wei Dong, Richard Socher, Li-Jia Li, Kai Li, and Li~Fei-Fei. 2009.
\newblock Imagenet: A large-scale hierarchical image database.
\newblock In \emph{2009 IEEE conference on computer vision and pattern recognition}, pages 248--255. Ieee.

\bibitem[{Ding et~al.(2023)Ding, Han, Xu, Zhang, and Li}]{Ding2023HiLMDTH}
Xinpeng Ding, Jianhua Han, Hang Xu, Wei Zhang, and X.~Li. 2023.
\newblock Hilm-d: Towards high-resolution understanding in multimodal large language models for autonomous driving.
\newblock \emph{ArXiv}, abs/2309.05186.

\bibitem[{Driess et~al.(2023)Driess, Xia, Sajjadi, Lynch, Chowdhery, Ichter, Wahid, Tompson, Vuong, Yu, Huang, Chebotar, Sermanet, Duckworth, Levine, Vanhoucke, Hausman, Toussaint, Greff, Zeng, Mordatch, and Florence}]{Driess2023PaLMEAE}
Danny Driess, F.~Xia, Mehdi S.~M. Sajjadi, Corey Lynch, Aakanksha Chowdhery, Brian Ichter, Ayzaan Wahid, Jonathan Tompson, Quan~Ho Vuong, Tianhe Yu, Wenlong Huang, Yevgen Chebotar, Pierre Sermanet, Daniel Duckworth, Sergey Levine, Vincent Vanhoucke, Karol Hausman, Marc Toussaint, Klaus Greff, Andy Zeng, Igor Mordatch, and Peter~R. Florence. 2023.
\newblock Palm-e: An embodied multimodal language model.
\newblock In \emph{International Conference on Machine Learning}.

\bibitem[{Fan et~al.(2024)Fan, You, Han, Liu, Tao, Huang, He, and Yang}]{fan2024vitar}
Qihang Fan, Quanzeng You, Xiaotian Han, Yongfei Liu, Yunzhe Tao, Huaibo Huang, Ran He, and Hongxia Yang. 2024.
\newblock Vitar: Vision transformer with any resolution.
\newblock \emph{arXiv preprint arXiv:2403.18361}.

\bibitem[{Goyal et~al.(2017)Goyal, Khot, Summers-Stay, Batra, and Parikh}]{VQAv2}
Yash Goyal, Tejas Khot, Douglas Summers-Stay, Dhruv Batra, and Devi Parikh. 2017.
\newblock Making the v in vqa matter: Elevating the role of image understanding in visual question answering.
\newblock In \emph{Proceedings of the IEEE conference on computer vision and pattern recognition}, pages 6904--6913.

\bibitem[{Guo et~al.(2025)Guo, Xu, Yao, Cui, Ni, Ge, Chua, Liu, and Huang}]{guo2025llava_uhd}
Zonghao Guo, Ruyi Xu, Yuan Yao, Junbo Cui, Zanlin Ni, Chunjiang Ge, Tat-Seng Chua, Zhiyuan Liu, and Gao Huang. 2025.
\newblock Llava-uhd: an lmm perceiving any aspect ratio and high-resolution images.
\newblock In \emph{European Conference on Computer Vision}, pages 390--406. Springer.

\bibitem[{Gurari et~al.(2018)Gurari, Li, Stangl, Guo, Lin, Grauman, Luo, and Bigham}]{vizwiz}
Danna Gurari, Qing Li, Abigale~J Stangl, Anhong Guo, Chi Lin, Kristen Grauman, Jiebo Luo, and Jeffrey~P Bigham. 2018.
\newblock Vizwiz grand challenge: Answering visual questions from blind people.
\newblock In \emph{Proceedings of the IEEE conference on computer vision and pattern recognition}, pages 3608--3617.

\bibitem[{Hong et~al.(2023)Hong, Wang, Lv, Xu, Yu, Ji, Wang, Wang, Dong, Ding, and Tang}]{Hong2023CogAgentAV}
Wenyi Hong, Weihan Wang, Qingsong Lv, Jiazheng Xu, Wenmeng Yu, Junhui Ji, Yan Wang, Zihan Wang, Yuxiao Dong, Ming Ding, and Jie Tang. 2023.
\newblock Cogagent: A visual language model for gui agents.
\newblock \emph{2024 IEEE/CVF Conference on Computer Vision and Pattern Recognition (CVPR)}, pages 14281--14290.

\bibitem[{Hu et~al.(2024)Hu, Xu, Ye, Yan, Zhang, Zhang, Li, Zhang, Jin, Huang, and Zhou}]{Hu2024mPLUGDocOwl1U}
Anwen Hu, Haiyang Xu, Jiabo Ye, Mingshi Yan, Liang Zhang, Bo~Zhang, Chen Li, Ji~Zhang, Qin Jin, Fei Huang, and Jingren Zhou. 2024.
\newblock mplug-docowl 1.5: Unified structure learning for ocr-free document understanding.
\newblock \emph{ArXiv}, abs/2403.12895.

\bibitem[{Hudson and Manning(2019)}]{hudson2019gqa}
Drew~A Hudson and Christopher~D Manning. 2019.
\newblock Gqa: A new dataset for real-world visual reasoning and compositional question answering.
\newblock In \emph{Proceedings of the IEEE/CVF conference on computer vision and pattern recognition}, pages 6700--6709.

\bibitem[{IDEFICS(2023)}]{idefics}
IDEFICS. 2023.
\newblock Introducing idefics: An open reproduction of state-of-the-art visual language model.
\newblock \url{https://huggingface.co/blog/idefics}.

\bibitem[{Khan et~al.(2022)Khan, Naqvi, and Meijering}]{khan2022leveraging}
Tariq~M Khan, Syed~S Naqvi, and Erik Meijering. 2022.
\newblock Leveraging image complexity in macro-level neural network design for medical image segmentation.
\newblock \emph{Scientific Reports}, 12(1):22286.

\bibitem[{Krishna et~al.(2017)Krishna, Zhu, Groth, Johnson, Hata, Kravitz, Chen, Kalantidis, Li, Shamma et~al.}]{krishna2017VisualGenome}
Ranjay Krishna, Yuke Zhu, Oliver Groth, Justin Johnson, Kenji Hata, Joshua Kravitz, Stephanie Chen, Yannis Kalantidis, Li-Jia Li, David~A Shamma, et~al. 2017.
\newblock Visual genome: Connecting language and vision using crowdsourced dense image annotations.
\newblock \emph{International journal of computer vision}, 123:32--73.

\bibitem[{Li et~al.(2023{\natexlab{a}})Li, Zhang, Chen, Wang, Yang, and Liu}]{Li2023OtterAM}
Bo~Li, Yuanhan Zhang, Liangyu Chen, Jinghao Wang, Jingkang Yang, and Ziwei Liu. 2023{\natexlab{a}}.
\newblock Otter: A multi-modal model with in-context instruction tuning.
\newblock \emph{ArXiv}, abs/2305.03726.

\bibitem[{Li et~al.(2023{\natexlab{b}})Li, Li, Savarese, and Hoi}]{blip2}
Junnan Li, Dongxu Li, Silvio Savarese, and Steven Hoi. 2023{\natexlab{b}}.
\newblock Blip-2: Bootstrapping language-image pre-training with frozen image encoders and large language models.
\newblock In \emph{International conference on machine learning}, pages 19730--19742. PMLR.

\bibitem[{Li et~al.(2024{\natexlab{a}})Li, Chen, Cai, Chen, Hong, Chen, Shen, and Gan}]{Li2024FlexAttentionFE}
Junyan Li, Delin Chen, Tianle Cai, Peihao Chen, Yining Hong, Zhenfang Chen, Yikang Shen, and Chuang Gan. 2024{\natexlab{a}}.
\newblock Flexattention for efficient high-resolution vision-language models.
\newblock \emph{ArXiv}, abs/2407.20228.

\bibitem[{Li et~al.(2024{\natexlab{b}})Li, Zhang, Wang, Zhong, Chen, Chu, Liu, and Jia}]{Li2024MiniGeminiMT}
Yanwei Li, Yuechen Zhang, Chengyao Wang, Zhisheng Zhong, Yixin Chen, Ruihang Chu, Shaoteng Liu, and Jiaya Jia. 2024{\natexlab{b}}.
\newblock Mini-gemini: Mining the potential of multi-modality vision language models.
\newblock \emph{ArXiv}, abs/2403.18814.

\bibitem[{Li et~al.(2023{\natexlab{c}})Li, Yang, Liu, Ma, Zhang, Yang, Sun, Liu, and Bai}]{Li2023MonkeyIR}
Zhang Li, Biao Yang, Qiang Liu, Zhiyin Ma, Shuo Zhang, Jingxu Yang, Yabo Sun, Yuliang Liu, and Xiang Bai. 2023{\natexlab{c}}.
\newblock Monkey: Image resolution and text label are important things for large multi-modal models.
\newblock \emph{2024 IEEE/CVF Conference on Computer Vision and Pattern Recognition (CVPR)}, pages 26753--26763.

\bibitem[{Lin et~al.(2014)Lin, Maire, Belongie, Hays, Perona, Ramanan, Doll{\'a}r, and Zitnick}]{coco}
Tsung-Yi Lin, Michael Maire, Serge Belongie, James Hays, Pietro Perona, Deva Ramanan, Piotr Doll{\'a}r, and C~Lawrence Zitnick. 2014.
\newblock Microsoft coco: Common objects in context.
\newblock In \emph{Computer Vision--ECCV 2014: 13th European Conference, Zurich, Switzerland, September 6-12, 2014, Proceedings, Part V 13}, pages 740--755. Springer.

\bibitem[{Liu et~al.(2023{\natexlab{a}})Liu, Li, Li, and Lee}]{liu2023improvedllava}
Haotian Liu, Chunyuan Li, Yuheng Li, and Yong~Jae Lee. 2023{\natexlab{a}}.
\newblock Improved baselines with visual instruction tuning.

\bibitem[{Liu et~al.(2024)Liu, Li, Li, Li, Zhang, Shen, and Lee}]{liu2024llavanext}
Haotian Liu, Chunyuan Li, Yuheng Li, Bo~Li, Yuanhan Zhang, Sheng Shen, and Yong~Jae Lee. 2024.
\newblock \href {https://llava-vl.github.io/blog/2024-01-30-llava-next/} {Llava-next: Improved reasoning, ocr, and world knowledge}.

\bibitem[{Liu et~al.(2023{\natexlab{b}})Liu, Li, Wu, and Lee}]{liu2023llava}
Haotian Liu, Chunyuan Li, Qingyang Wu, and Yong~Jae Lee. 2023{\natexlab{b}}.
\newblock Visual instruction tuning.

\bibitem[{Liu et~al.(2023{\natexlab{c}})Liu, Duan, Zhang, Li, Zhang, Zhao, Yuan, Wang, He, Liu et~al.}]{liu2023mmbench}
Yuan Liu, Haodong Duan, Yuanhan Zhang, Bo~Li, Songyang Zhang, Wangbo Zhao, Yike Yuan, Jiaqi Wang, Conghui He, Ziwei Liu, et~al. 2023{\natexlab{c}}.
\newblock Mmbench: Is your multi-modal model an all-around player?
\newblock \emph{arXiv preprint arXiv:2307.06281}.

\bibitem[{Lu et~al.(2022)Lu, Mishra, Xia, Qiu, Chang, Zhu, Tafjord, Clark, and Kalyan}]{scienceqa}
Pan Lu, Swaroop Mishra, Tanglin Xia, Liang Qiu, Kai-Wei Chang, Song-Chun Zhu, Oyvind Tafjord, Peter Clark, and Ashwin Kalyan. 2022.
\newblock Learn to explain: Multimodal reasoning via thought chains for science question answering.
\newblock \emph{Advances in Neural Information Processing Systems}, 35:2507--2521.

\bibitem[{Lv et~al.(2023)Lv, Huang, Chen, Cui, Ma, Chang, Huang, Wang, Dong, Luo, Wu, Wang, Zhang, and Wei}]{Lv2023Kosmos25AM}
Tengchao Lv, Yupan Huang, Jingye Chen, Lei Cui, Shuming Ma, Ya-Chi Chang, Shaohan Huang, Wenhui Wang, Li~Dong, Weiyao Luo, Shaoxiang Wu, Guoxin Wang, Cha Zhang, and Furu Wei. 2023.
\newblock Kosmos-2.5: A multimodal literate model.
\newblock \emph{ArXiv}, abs/2309.11419.

\bibitem[{Machado et~al.(2015)Machado, Romero, Nadal, Santos, Correia, and Carballal}]{machado2015computerized}
Penousal Machado, Juan Romero, Marcos Nadal, Antonino Santos, Jo{\~a}o Correia, and Adri{\'a}n Carballal. 2015.
\newblock Computerized measures of visual complexity.
\newblock \emph{Acta psychologica}, 160:43--57.

\bibitem[{Mahon and Lukasiewicz(2023)}]{image_complexity}
Louis Mahon and Thomas Lukasiewicz. 2023.
\newblock Minimum description length clustering to measure meaningful image complexity.
\newblock \emph{arXiv preprint arXiv:2306.14937}.

\bibitem[{Marino et~al.(2019)Marino, Rastegari, Farhadi, and Mottaghi}]{okvqa}
Kenneth Marino, Mohammad Rastegari, Ali Farhadi, and Roozbeh Mottaghi. 2019.
\newblock Ok-vqa: A visual question answering benchmark requiring external knowledge.
\newblock In \emph{Proceedings of the IEEE/cvf conference on computer vision and pattern recognition}, pages 3195--3204.

\bibitem[{McKinzie et~al.(2024{\natexlab{a}})McKinzie, Gan, Fauconnier, Dodge, Zhang, Dufter, Shah, Du, Peng, Weers et~al.}]{mckinzie2024mm1}
Brandon McKinzie, Zhe Gan, Jean-Philippe Fauconnier, Sam Dodge, Bowen Zhang, Philipp Dufter, Dhruti Shah, Xianzhi Du, Futang Peng, Floris Weers, et~al. 2024{\natexlab{a}}.
\newblock Mm1: Methods, analysis \& insights from multimodal llm pre-training.
\newblock \emph{arXiv preprint arXiv:2403.09611}.

\bibitem[{McKinzie et~al.(2024{\natexlab{b}})McKinzie, Gan, Fauconnier, Dodge, Zhang, Dufter, Shah, Du, Peng, Weers et~al.}]{mm1}
Brandon McKinzie, Zhe Gan, Jean-Philippe Fauconnier, Sam Dodge, Bowen Zhang, Philipp Dufter, Dhruti Shah, Xianzhi Du, Futang Peng, Floris Weers, et~al. 2024{\natexlab{b}}.
\newblock Mm1: Methods, analysis \& insights from multimodal llm pre-training.
\newblock \emph{arXiv preprint arXiv:2403.09611}.

\bibitem[{Mishra et~al.(2019)Mishra, Shekhar, Singh, and Chakraborty}]{mishra2019ocrvqa}
Anand Mishra, Shashank Shekhar, Ajeet~Kumar Singh, and Anirban Chakraborty. 2019.
\newblock Ocr-vqa: Visual question answering by reading text in images.
\newblock In \emph{2019 international conference on document analysis and recognition (ICDAR)}, pages 947--952. IEEE.

\bibitem[{Raghu et~al.(2021)Raghu, Unterthiner, Kornblith, Zhang, and Dosovitskiy}]{raghu2021vision}
Maithra Raghu, Thomas Unterthiner, Simon Kornblith, Chiyuan Zhang, and Alexey Dosovitskiy. 2021.
\newblock Do vision transformers see like convolutional neural networks?
\newblock \emph{Advances in neural information processing systems}, 34:12116--12128.

\bibitem[{Redies et~al.(2012)Redies, Amirshahi, Koch, and Denzler}]{redies2012phog}
Christoph Redies, Seyed~Ali Amirshahi, Michael Koch, and Joachim Denzler. 2012.
\newblock Phog-derived aesthetic measures applied to color photographs of artworks, natural scenes and objects.
\newblock In \emph{European conference on computer vision}, pages 522--531. Springer.

\bibitem[{Sabottke and Spieler(2020)}]{sabottke2020effect}
Carl~F Sabottke and Bradley~M Spieler. 2020.
\newblock The effect of image resolution on deep learning in radiography.
\newblock \emph{Radiology: Artificial Intelligence}, 2(1):e190015.

\bibitem[{Singh et~al.(2019)Singh, Natarajan, Shah, Jiang, Chen, Batra, Parikh, and Rohrbach}]{textvqa}
Amanpreet Singh, Vivek Natarajan, Meet Shah, Yu~Jiang, Xinlei Chen, Dhruv Batra, Devi Parikh, and Marcus Rohrbach. 2019.
\newblock Towards vqa models that can read.
\newblock In \emph{Proceedings of the IEEE/CVF conference on computer vision and pattern recognition}, pages 8317--8326.

\bibitem[{Team(2024)}]{Team2024ChameleonME}
Chameleon Team. 2024.
\newblock Chameleon: Mixed-modal early-fusion foundation models.
\newblock \emph{ArXiv}, abs/2405.09818.

\bibitem[{Tian et~al.(2023)Tian, Wu, Dai, Hu, Qiao, and Jiang}]{tian2023resformer}
Rui Tian, Zuxuan Wu, Qi~Dai, Han Hu, Yu~Qiao, and Yu-Gang Jiang. 2023.
\newblock Resformer: Scaling vits with multi-resolution training.
\newblock In \emph{Proceedings of the IEEE/CVF Conference on Computer Vision and Pattern Recognition}, pages 22721--22731.

\bibitem[{Tong et~al.(2024)Tong, Brown, Wu, Woo, Middepogu, Akula, Yang, Yang, Iyer, Pan et~al.}]{tong2024cambrian}
Shengbang Tong, Ellis Brown, Penghao Wu, Sanghyun Woo, Manoj Middepogu, Sai~Charitha Akula, Jihan Yang, Shusheng Yang, Adithya Iyer, Xichen Pan, et~al. 2024.
\newblock Cambrian-1: A fully open, vision-centric exploration of multimodal llms.
\newblock \emph{arXiv preprint arXiv:2406.16860}.

\bibitem[{Wang et~al.(2024)Wang, Bai, Tan, Wang, Fan, Bai, Chen, Liu, Wang, Ge et~al.}]{wang2024qwen2vl}
Peng Wang, Shuai Bai, Sinan Tan, Shijie Wang, Zhihao Fan, Jinze Bai, Keqin Chen, Xuejing Liu, Jialin Wang, Wenbin Ge, et~al. 2024.
\newblock Qwen2-vl: Enhancing vision-language model's perception of the world at any resolution.
\newblock \emph{arXiv preprint arXiv:2409.12191}.

\bibitem[{Wei et~al.(2023)Wei, Kong, Chen, Zhao, Ge, Yang, Sun, Han, and Zhang}]{Wei2023VarySU}
Haoran Wei, Lingyu Kong, Jinyue Chen, Liang Zhao, Zheng Ge, Jinrong Yang, Jian‐Yuan Sun, Chunrui Han, and Xiangyu Zhang. 2023.
\newblock Vary: Scaling up the vision vocabulary for large vision-language models.
\newblock \emph{ArXiv}, abs/2312.06109.

\bibitem[{wen Dong et~al.(2024)wen Dong, Zhang, Zang, Cao, Wang, Ouyang, Zhang, Duan, Zhang, Li, Yan, Gao, Chen, Zhang, Li, Li, Wang, Chen, He, Zhang, Dai, Qiao, Lin, and Wang}]{Dong2024InternLMXComposer24KHDAP}
Xiao wen Dong, Pan Zhang, Yuhang Zang, Yuhang Cao, Bin Wang, Linke Ouyang, Songyang Zhang, Haodong Duan, Wenwei Zhang, Yining Li, Hang Yan, Yang Gao, Zhe Chen, Xinyue Zhang, Wei Li, Jingwen Li, Wenhai Wang, Kai Chen, Conghui He, Xingcheng Zhang, Jifeng Dai, Yuxin Qiao, Dahua Lin, and Jiaqi Wang. 2024.
\newblock Internlm-xcomposer2-4khd: A pioneering large vision-language model handling resolutions from 336 pixels to 4k hd.
\newblock \emph{ArXiv}, abs/2404.06512.

\bibitem[{Wu and Xie(2023)}]{Wu2023VGV}
Penghao Wu and Saining Xie. 2023.
\newblock \href {https://api.semanticscholar.org/CorpusID:266436019} {V*: Guided visual search as a core mechanism in multimodal llms}.
\newblock \emph{2024 IEEE/CVF Conference on Computer Vision and Pattern Recognition (CVPR)}, pages 13084--13094.

\bibitem[{Xue et~al.(2024)Xue, Shu, Awadalla, Wang, Yan, Purushwalkam, Zhou, Prabhu, Dai, Ryoo et~al.}]{xue2024xgen}
Le~Xue, Manli Shu, Anas Awadalla, Jun Wang, An~Yan, Senthil Purushwalkam, Honglu Zhou, Viraj Prabhu, Yutong Dai, Michael~S Ryoo, et~al. 2024.
\newblock xgen-mm (blip-3): A family of open large multimodal models.
\newblock \emph{arXiv preprint arXiv:2408.08872}.

\bibitem[{Yao et~al.(2024)Yao, Yu, Zhang, Wang, Cui, Zhu, Cai, Li, Zhao, He et~al.}]{yao2024minicpm_v}
Yuan Yao, Tianyu Yu, Ao~Zhang, Chongyi Wang, Junbo Cui, Hongji Zhu, Tianchi Cai, Haoyu Li, Weilin Zhao, Zhihui He, et~al. 2024.
\newblock Minicpm-v: A gpt-4v level mllm on your phone.
\newblock \emph{arXiv preprint arXiv:2408.01800}.

\bibitem[{Yin et~al.(2023)Yin, Fu, Zhao, Li, Sun, Xu, and Chen}]{Yin2023ASO}
Shukang Yin, Chaoyou Fu, Sirui Zhao, Ke~Li, Xing Sun, Tong Xu, and Enhong Chen. 2023.
\newblock A survey on multimodal large language models.
\newblock \emph{ArXiv}, abs/2306.13549.

\bibitem[{Zhang et~al.(2023)Zhang, Khayatkhoei, Chhikara, and Ilievski}]{Zhang2023TowardsPS}
Jiarui Zhang, Mahyar Khayatkhoei, Prateek Chhikara, and Filip Ilievski. 2023.
\newblock \href {https://api.semanticscholar.org/CorpusID:264439597} {Towards perceiving small visual details in zero-shot visual question answering with multimodal llms}.

\bibitem[{Zhang et~al.(2024)Zhang, wen Dong, Zang, Cao, Qian, Chen, Guo, Duan, Wang, Ouyang, Zhang, Zhang, Li, Gao, Sun, Zhang, Li, Li, Wang, Yan, He, Zhang, Chen, Dai, Qiao, Lin, and Wang}]{Zhang2024InternLMXComposer25AV}
Pan Zhang, Xiao wen Dong, Yuhang Zang, Yuhang Cao, Rui Qian, Lin Chen, Qipeng Guo, Haodong Duan, Bin Wang, Linke Ouyang, Songyang Zhang, Wenwei Zhang, Yining Li, Yang Gao, Peng Sun, Xinyue Zhang, Wei Li, Jingwen Li, Wenhai Wang, Hang Yan, Conghui He, Xingcheng Zhang, Kai Chen, Jifeng Dai, Yu~Qiao, Dahua Lin, and Jiaqi Wang. 2024.
\newblock Internlm-xcomposer-2.5: A versatile large vision language model supporting long-contextual input and output.
\newblock \emph{ArXiv}, abs/2407.03320.

\bibitem[{Zhao et~al.(2024)Zhao, Li, Duan, Huang, Li, Chen, and Yang}]{Zhao2024MGLLaVATM}
Xiangyu Zhao, Xiangtai Li, Haodong Duan, Haian Huang, Yining Li, Kai Chen, and Hua Yang. 2024.
\newblock \href {https://api.semanticscholar.org/CorpusID:270710948} {Mg-llava: Towards multi-granularity visual instruction tuning}.
\newblock \emph{ArXiv}, abs/2406.17770.

\bibitem[{Zhou et~al.(2023)Zhou, Liu, Yurtsever, Žagar, Zimmer, Cao, and Knoll}]{Zhou2023VisionLM}
Xingcheng Zhou, Mingyu Liu, Ekim Yurtsever, Bare~Luka Žagar, Walter Zimmer, Hu~Cao, and Alois~C. Knoll. 2023.
\newblock Vision language models in autonomous driving: A survey and outlook.
\newblock \emph{IEEE Transactions on Intelligent Vehicles}.

\end{thebibliography}

\appendix

\section{Dynamic Resolution and High-Resolution Techniques in VLLMs}
\label{appendix:sec:dynamic_high_res_vllms}

\paragraph{Native Dynamic Resolution VLLMs.}
A significant line of research focuses on VLLMs with native capabilities to handle dynamic input resolutions, often through architectural innovations or specialized pre-training. For instance, \textbf{Qwen2VL}~\citep{wang2024qwen2vl} employs 2D RoPE for flexible positional encoding. \textbf{MiniCPM-V}~\citep{yao2024minicpm_v} focuses on efficient high-resolution processing, sometimes using multi-scale vision encoders. \textbf{LLaVA-UHD}~\citep{guo2025llava_uhd} introduces strategies for ultra-high-definition images and varied aspect ratios, often involving intelligent image slicing. \textbf{InternLM-XComposer2-4KHD}~\citep{Dong2024InternLMXComposer24KHDAP} also demonstrates strong capabilities in handling very high resolutions through sophisticated tiling strategies. While these models offer great flexibility, they typically require substantial pre-training and may not explicitly optimize for a single best resolution per task. Our approach, in contrast, focuses on lightweight, post-hoc adaptation of existing VLLMs to a task-specific optimal resolution.

\paragraph{Other High-Resolution Processing Techniques.}
Beyond models with end-to-end dynamic resolution, other techniques enable VLLMs to process high-resolution information. Some works focus on \textbf{using or adapting vision encoders} to directly support higher resolutions within a VLLM framework, such as CogAgent~\citep{Hong2023CogAgentAV} with its dense feature integration, or models like MiniGemini~\citep{Li2024MiniGeminiMT}, Kosmos-2.5~\citep{Lv2023Kosmos25AM}, and Vary~\citep{Wei2023VarySU}.
\textbf{Patchification and tiling strategies} are common, where high-resolution images are divided into smaller patches processed by standard encoders, with subsequent feature aggregation; examples include Monkey~\citep{Li2023MonkeyIR}, mPLUG-DocOwl~\citep{Hu2024mPLUGDocOwl1U}, and LLaVA-NEXT~\citep{liu2024llavanext}.
\textbf{Region-aware processing} aims to focus on salient regions, with methods like V*~\citep{Wu2023VGV} selecting relevant regions for fine-grained understanding, MG-LLaVA~\citep{Zhao2024MGLLaVATM} using multi-grained GNNs, and PS-VLLM~\citep{Zhang2023TowardsPS} progressively selecting visual tokens.
To optimize computational costs associated with high resolutions, FlexAttention~\citep{Li2024FlexAttentionFE} employs dual tokenization for selective processing of high-resolution tokens.

Our work complements these techniques by first providing a mechanism to determine a task-optimal discrete resolution, to which a model (potentially employing some of these techniques) can then be adapted.

\section{More Implementation Details}
\label{sec:appendix-implementation-details}

\subsection{Vision-Language Tasks}
\label{subsec:appendix-vision-language-tasks}
\textit{Science-QA}~\cite{scienceqa}, a multimodal science question answering benchmark featuring over 21k multiple-choice questions on diverse topics. The visual component includes natural images and diagrams, testing the model's ability to integrate both textual and visual information for coherent reasoning and explanation generation.
\textit{Vizwiz}~\cite{vizwiz}, a dataset derived from real-world images paired with spoken questions from visually impaired individuals. This task assesses a model’s ability to process low-quality, unstructured images and generate accurate responses to conversational queries.
\textit{VQAv2}~\cite{VQAv2}, an expanded version of the original Visual Question Answering (VQA) dataset, designed to reduce language biases. It challenges models to deeply understand visual content in order to answer questions about pairs of semantically similar yet visually distinct images.
\textit{TextVQA}~\cite{textvqa}, a dataset focusing on a model’s capacity to read and reason about textual elements in images, evaluating its ability to integrate Optical Character Recognition (OCR) with visual reasoning to answer questions.
\textit{OKVQA}~\cite{okvqa}, a benchmark that requires models to leverage external knowledge beyond image and question analysis, necessitating access to and reasoning with unstructured knowledge sources for accurate answers.
\textit{GQA}~\cite{hudson2019gqa}, a dataset designed for real-world visual reasoning and compositional question answering, requiring models to demonstrate strong multi-modal understanding, logical reasoning, and the ability to answer questions that necessitate connecting information across both visual and linguistic domains.
\textit{MMBench}~\cite{liu2023mmbench}, a comprehensive multimodal evaluation set with over 2,974 multiple-choice questions across 20 ability dimensions, providing a robust assessment of various vision-language skills, such as reasoning, comprehension, and explanation generation.
\textit{MMBench-CN}, a variant of MMBench focusing on tasks involving Chinese text and images, evaluating the model's proficiency in processing and understanding multilingual data.

\subsection{Baseline Methods}
\label{subsec:appendix-baseline-methods}
In addition to the original LLaVA model, we compare our method with several state-of-the-art approaches, including BLIP-2~\cite{blip2}, InstructBLIP~\citep{instructblip} (with LLM backbones at two scales), Shikra~\citep{chen2023shikra}, and IDEFICS~\citep{idefics} (also with LLM backbones at two scales), as well as Qwen-VL and Qwen-VL-Chat~\cite{qwen_vl}. The results for these baseline methods, along with LLaVA with the Vicuna-13B backbone, are cited from previous work~\citep{liu2023improvedllava}. For LLaVA with a Vicuna-7B backbone, we report our reproduced results across different vision-language tasks.

As a training-free baseline to extend the image input resolution, we apply positional embedding interpolation to extend the position embeddings of the vision encoder in LLaVA. This technique, widely used for Vision Transformers in VLLMs~\cite{qwen_vl, blip2}, allows models to handle higher image input resolutions than their original training resolution. We evaluate the performance of this extension without any additional training of the projector and the LLM backbone.

\subsection{Method details}
\label{subsec:appendix-method}

\paragraph{Image Complexity Heuristic Approach}
Image complexity for vision-language tasks is calculated using an open-source tool\footnote{https://github.com/Lou1sM/\\meaningful\_image\_complexity}. We utilize the author-recommended hyperparameters: the number of clusters is set to $8$, and the subsample rate is $0.8$. To reduce computational overhead, the input image resolution is set to $112 \times 112$, and two cluster levels are used, with their combined scores yielding the final complexity value. The complexity scores are normalized via min-max scaling, where the minimum and maximum values are computed from 100 sampled images from the ImageNet dataset~\cite{deng2009imagenet}.

\paragraph{RandAugment Perturbation on Image Input}
When assessing model variance across different resolutions, we apply random perturbations to each input image using the RandAugment algorithm, implemented via an existing tool\footnote{https://github.com/TorchSSL/TorchSSL/blob/main/\\datasets/augmentation/randaugment.py}. For each image, we perform three random augmentations. To mitigate the effects of randomness and enhance result stability, we repeat the variance measurement process three times, each using a different random seed. The final uncertainty variance is obtained by averaging the results from these three iterations.

\subsection{More Parameter-Efficient Fine-Tuning Details}
\label{subsec:appendix-PEFT}
The standard training hyperparameters are largely preserved, as outlined in Table~\ref{tab:training-hyperparameters}, with two notable adjustments for image resolutions of $560^2$ and $672^2$: (1) The learning rate is reduced from $2e-5$ to $1e-5$ to prevent training loss explosion observed with the original rate. (2) The maximum number of tokens is increased from $2048$ to $3072$ and $4096$, respectively, to accommodate the increased number of image tokens.

Post-training experiments are conducted on eight NVIDIA GeForce RTX 4090 GPUs, with training time costs detailed in Table~\ref{tab:training-time-cost}. Due to GPU memory limitations, DeepSpeed ZeRO-3 was employed for training at the resolution of $672^2$, while ZeRO-2 was used for other resolutions. This accounts for the significant increase in training time between $672^2$ and $560^2$.

In the ablation study (Section~\ref{subsec:ablation-study}), we separately fine-tune only the projector and only the position embeddings, using the stage 1 setting for consistency with the goals of the different training stages. The corresponding hyperparameters are also detailed in Table~\ref{tab:training-hyperparameters}.

\begin{table}
\centering
\caption{Hyperparameters at two training stages}
\resizebox{\linewidth}{!}{\begin{tabular}{l|ccccccc}
\toprule
Hyperparameter & batch size & lr   & lr schedule                    & weight decay       & epoch              & optimizer              & max tokens             \\
\midrule
Stage 1        & 256        & 1e-3 & \multirow{2}{*}{cosinie decay} & \multirow{2}{*}{0} & \multirow{2}{*}{1} & \multirow{2}{*}{AdamW} & \multirow{2}{*}{2048}  \\
Stage 2        & 128        & 2e-4 &                                &                    &                    &                        &                       \\
\bottomrule
\end{tabular}}
\label{tab:training-hyperparameters}
\end{table}

\begin{table}
\centering
\caption{Training time cost}
\resizebox{\linewidth}{!}{\begin{tabular}{l|ccccc}
\toprule
Resolution         & $224\times 224$     & $336\times 336$     & $448\times 448$    & $560\times560$       & $672\times 672$       \\
\midrule
Training Time Cost & 11h 50m & 16h 17m & 24h 7m & 32h 29m & 124h 44m \\
\bottomrule
\end{tabular}}
\label{tab:training-time-cost}
\end{table}

\section{Impact of Statistical Distributions on Empirical Formula Performance}
\label{appendix:sec-stat-impact-on-formula}

To evaluate the extent to which the statistical distributions of complexity $C(T)$ and uncertainty variance $V(T)$ influence the performance of the empirical formula, we present the standard deviations of $C(T)$ and $V(T)$  for each vision-language task, along with their respective ratios to the mean values. These statistics are detailed in Table~\ref{tab:stat-dist-ct-vt}.

\begin{table}
\centering
\caption{Statistical characteristics of $C(T)$ and $V(T)$ in each task. SD represents Standard Deviation, and Ratio indicates the ratio of the standard deviation to the mean.}
\resizebox{\linewidth}{!}{
\begin{tabular}{lcccc}
\toprule
Task          & $C(T)$ SD & $C(T)$ Ratio & $V(T)$ SD & $V(T)$ Ratio  \\
\midrule
ScienceQA-IMG & 3.3633  & 0.2384     & 0.4398  & 2.5466      \\
\midrule
Vizwiz        & 2.4405  & 0.1541     & 0.3383  & 6.0196      \\
\midrule
VQAv2         & 2.2005  & 0.1242     & 0.7925  & 4.2562      \\
\midrule
GQA           & 1.6582  & 0.0910     & 1.2595  & 4.9103      \\
\midrule
TextVQA       & 2.3057  & 0.1318     & 0.5258  & 3.3405      \\
\midrule
OKVQA         & 2.1958  & 0.1224     & 0.5487  & 3.7711      \\
\midrule
MMBench       & 3.5426  & 0.2196     & 1.2040  & 2.8915      \\
\midrule
MMBench-CN    & 3.5482  & 0.2197     & 1.0840  & 2.8310 \\
\bottomrule
\end{tabular}
}
\label{tab:stat-dist-ct-vt}
\end{table}

\begin{figure}
    \centering     
    \includegraphics[width=0.8\linewidth]{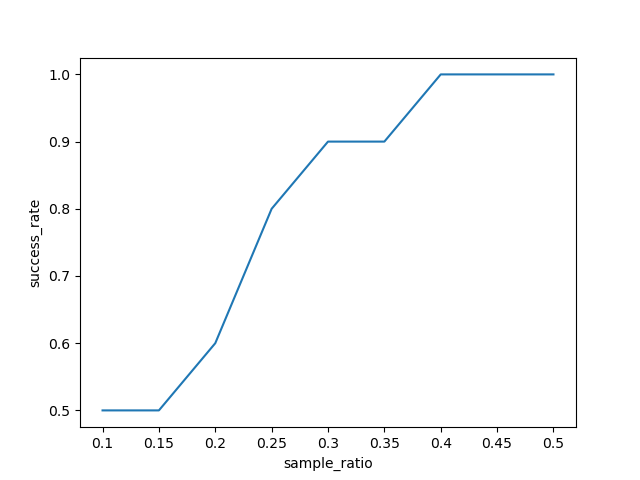}
    \caption{Relationship between sampling ratio and the success rate of the empirical formula.}
    \label{fig:sample-ratio-succ-rate}
\end{figure}

The results indicate that $C(T)$ exhibits relatively low variance across tasks, whereas $V(T)$ shows substantially higher variability. This observation justifies our decision to adopt task-wise selection instead of sample-wise selection, as the higher variability in $V(T)$ at the sample level complicates consistent prediction.

To further assess the influence of $C(T)$ and $V(T)$ variance on the effectiveness of the empirical formula, we conducted an additional experiment. Specifically, we randomly sampled subsets of varying proportions from the original dataset and computed the average $C(T)$ and $V(T)$ values for these subsets to estimate task-level statistics. We then evaluated the empirical formula, previously tuned using a hyperparameter $k$ on three reference tasks, to predict the optimal resolution across all tasks under these conditions.

The sampling proportions vary from 10\% to 50\%, with each experiment repeated 10 times using different random seeds. The success rate was defined as the percentage of instances where the empirical formula accurately predicted the optimal resolution for all tasks. The results, presented in Figure~\ref{fig:sample-ratio-succ-rate}, reveal the following key findings: (1) At a sampling ratio of 40\%, the success rate reaches 100\%, demonstrating the empirical formula's robustness in predicting the optimal resolution. (2) At a sampling ratio of 10\%, the success rate drops to 50\%, indicating that a smaller subset size introduces variability that adversely affects prediction accuracy.

These findings highlight that while reducing the dataset size can lower computational costs, excessively small subsets may lead to suboptimal predictions. Moreover, the current approach relies on random sampling; future exploration of more advanced sampling strategies that select representative samples could potentially achieve high success rates with smaller subsets.

\section{Acknowledgment of AI Assistance in Writing and Revision}
We utilized LLMs for revising and enhancing writing of this paper.

\end{document}